\pdfoutput=1
\documentclass{article}

\usepackage[main,final,nonatbib]{neurips_2026}
\usepackage[numbers]{natbib}

\usepackage{amsmath,amssymb,amsthm}
\usepackage{booktabs}
\usepackage{graphicx}
\usepackage{hyperref}
\hypersetup{
  pdftitle={Derivative-Informed Training of Neural Operators On-the-Fly via Sketched Tangent Consistency},
  pdfauthor={Xinhan Yang, Lu Lu, Shancong Mou}
}
\usepackage{cleveref}
\usepackage{algorithm,algorithmic}
\usepackage{enumitem}
\usepackage{xcolor}
\usepackage{subcaption}
\usepackage{float}
\usepackage{multirow}

\newcommand{\msdexp}[3]{(#1\mathbin{\pm}#2)\!\times\!10^{#3}}
\newcommand{\apptablesetup}{%
  \centering\small
  \setlength{\tabcolsep}{4pt}%
  \renewcommand{\arraystretch}{1.08}%
}

\newtheorem{proposition}{Proposition}

\newtheorem{remark}{Remark}

\title{ Derivative-Informed Training of Neural Operators On-the-Fly via Sketched Tangent Consistency}

\author{
  Xinhan Yang$^{1}$ \quad Lu Lu$^{2}$ \quad Shancong Mou$^{1}$ \\
  $^{1}$Department of Industrial and Systems Engineering,
  University of Minnesota \\
  Minneapolis, MN 55455 \\
  $^{2}$Department of Statistics and Data Science, Yale University \\
  New Haven, CT 06511, USA
}

\begin{document}
\raggedbottom
\maketitle

\begin{abstract}
Derivative-informed training improves neural operators by directly supervising their input-output sensitivities, which is crucial when neural operators are used as differentiable surrogates for inverse problems, PDE-constrained optimization, design, and control. However, existing methods rely on offline-generated derivative labels, making data generation slow, storage-intensive, and difficult to adapt across datasets, resolutions, or perturbation bases. We propose sketched tangent consistency loss (sTCL), an on-the-fly derivative-informed training objective that, for PDEs with a known and differentiable residual, enforces sensitivity consistency directly from the governing equation without offline tangent labels or neural-operator architecture changes. sTCL uses randomly sketched input perturbations to provide a lightweight derivative-level physics constraint during training. However, raw forward-sensitivity residual penalties can fail for stiff, ill-conditioned, indefinite, or coupled saddle-point tangent operators. To address this, we introduce lightweight operator-aware loss-conditioning mechanisms selected by a simple tangent-operator decision rule. Across Helmholtz, nonlinear diffusion--reaction, Burgers, Allen--Cahn, and Navier--Stokes, with the same neural-operator backbone for all methods, the PDE-specific sTCL losses achieve solution and Jacobian accuracy comparable to offline derivative-informed training (DIFNO) while eliminating the offline derivative-data generation and storage pipeline. These results show that on-the-fly derivative-informed training need not merely amortize offline tangent-solve cost into training; with appropriate sketching and loss design, sTCL provides an effective drop-in path to derivative-informed neural operators.
Code is available at \url{https://github.com/Mou-group-UMN/Derivative-informed-training-on-the-fly}.
\end{abstract}

\section{Introduction}
\label{sec:intro}

Neural operators~\cite{li2021fourier,lu2021learning,wang2021learning,kovachki2023neural} learn solution maps between function spaces. Given a parametric partial differential equation (PDE), a neural operator aims to approximate the solution map from an input function to the corresponding PDE solution. Beyond predicting the solution itself, trained neural operators are increasingly used as differentiable surrogates in downstream tasks such as PDE-constrained optimization, inverse problems, design, and control~\cite{lu2021physics,zhu2023fourier}. In these settings, the surrogate's derivative with respect to the input is often as important as its solution accuracy. For example, gradients of reduced objectives depend directly on the Jacobian of the solution operator. Thus, a neural operator that is accurate in solution value but inaccurate in its Jacobian may still produce unreliable optimization or control directions.

Derivative-informed, or Sobolev-style, training~\cite{czarnecki2017sobolev} addresses this issue by directly training the surrogate Jacobian. Recent methods such as DINO~\cite{o2024derivative}, DIFNO~\cite{yao2025difno}, and sensitivity-constrained FNO~\cite{behroozi2025sensitivity} train neural operators to match Jacobian--vector products of the true PDE solution operator along selected input perturbation directions. This additional derivative information has been shown to improve both solution accuracy and sensitivity accuracy, especially for nonlinear PDEs where standard mean-squared-error training provides little direct control over the learned Jacobian. However, existing derivative-informed neural-operator methods usually rely on offline-generated sensitivity labels~\cite{o2024derivative, yao2025difno, behroozi2025sensitivity}. For each training input and each selected perturbation direction, a classical PDE solver is used to solve the corresponding forward (or adjoint) sensitivity equation. These tangent solutions are then stored and used as derivative-supervision data during training. This pipeline is accurate, but it is also expensive and inflexible. In practice, each training sample requires one forward solve plus many linearized tangent solves, each typically comparable in cost to the forward solve, and the resulting derivative tensors must be stored on disk for use during training~\cite{yao2025difno}. For nonlinear PDEs at moderate dataset and discretization sizes, this offline stage can easily reach hours to days of computing time and tens of GB of disk storage per training set. Moreover, whenever the training distribution, dataset size, resolution, or derivative sketch basis changes, the sensitivity data must be regenerated from scratch. As a result, the offline tangent-solve stage becomes a major practical bottleneck for scalable and flexible derivative-informed training.

\textit{This paper asks whether, for PDEs whose governing equations are known, derivative-informed neural-operator training can be used \emph{on the fly} as a drop-in loss: without offline tangent solves, without stored derivative labels, without changing the neural-operator architecture, and without materializing the full surrogate Jacobian.} We answer yes through a systematic study that identifies two key ingredients:

\textbf{\emph{Sketching} makes the on-the-fly derivative penalty computationally feasible.} We introduce a \emph{sketched tangent consistency loss} (sTCL), which evaluates derivative consistency by randomly sampling perturbation directions in the input function space. At each training step, sTCL samples a small number of directions, computes the corresponding surrogate Jacobian--vector products using forward-mode automatic differentiation, and applies a PDE-derived consistency loss for the forward sensitivity equation. This avoids both full-Jacobian materialization and offline sensitivity-label generation. Computationally, the derivative penalty requires only a few Jacobian--vector products and a few applications of the known tangent operator, making its per-step cost the same order as the standard data-fitting loss. As a result, sTCL can be used as a lightweight regularization term on top of ordinary neural-operator training, rather than as a separate architecture or preprocessing pipeline. This stochastic directional construction mirrors the logic of \textit{physics-informed} neural networks/neural operators~\cite{raissi2019physics,wang2021learning}. In physics-informed training, one randomly samples space-time collocation points and enforces the governing PDE residual on average. In sTCL, one instead randomly samples directions in the input function space and enforces the forward sensitivity equation on average. Thus, sTCL can be viewed as a physics-informed training principle lifted from solution values to solution-operator derivatives.

\textbf{\emph{Conditioning} makes the resulting physics-informed sensitivity signal effective.} Although sketching makes on-the-fly derivative-informed training computationally tractable, the quality of the resulting training signal depends critically on the conditioning of the sensitivity residual. The forward-sensitivity residual is governed by the tangent operator associated with the PDE. If this operator is stiff, ill-conditioned, or indefinite, the raw residual loss can provide poor optimization signals. In fact, the conditioning of the residual loss can be substantially worse than that of the tangent equation itself. This issue is closely related to known optimization pathologies of physics-informed residual losses~\cite{wang2021understanding,wang2022when,krishnapriyan2021characterizing} and recent operator-preconditioning interpretations of PINN training difficulty~\cite{de2024operator}. These observations explain why naive derivative-label-free sensitivity penalties may fail even when they are cheap to evaluate. We therefore study operator conditioning as a central design principle for on-the-fly derivative-informed training. We introduce lightweight operator-aware loss-conditioning mechanisms selected according to the structure of the tangent operator. For symmetric positive-definite elliptic operators, we use a DST-based inverse-Laplacian residual loss. For symmetric indefinite operators, where residual minimization can fail near resonance modes, we use a short-iteration MINRES target-matching form with a shifted-Laplacian preconditioner. For well-conditioned parabolic operators, a raw or lightly preconditioned residual is often sufficient. These mechanisms are inexpensive and preserve the on-the-fly nature of the method.

To demonstrate that on-the-fly derivative-informed training can replace offline derivative labels in practice, we conduct a study across five PDE benchmarks: Helmholtz, nonlinear diffusion--reaction, Burgers, Allen--Cahn, and Navier--Stokes. For each PDE, we evaluate three data regimes, yielding $15$ comparison cells. We use a controlled training protocol with shared architectures, optimizers, batch sizes, and matched direction sampling; Helmholtz, nonlinear diffusion--reaction, and Burgers are trained for $2000$ epochs, Allen--Cahn is trained for $1500$ epochs, and Navier--Stokes is trained for $5000$ epochs. Loss weights are selected on validation grids for all five benchmarks, as detailed in \cref{app:per_cell_configs}. The central comparison is sTCL versus DIFNO~\cite{yao2025difno}, the closest offline derivative-supervision baseline, under the same FNO backbone, data, and training protocol, so that differences reflect the derivative-supervision strategy rather than the architecture; sTCL itself is architecture-agnostic. Across both solution-error and Jacobian-error evaluations, sTCL achieves accuracy comparable to DIFNO while eliminating offline tangent-data generation and derivative-label storage. Its training time is also comparable to that of DIFNO (\cref{tab:wallclock_per_pde}), even before accounting for DIFNO's offline tangent-label generation, which sTCL avoids entirely. This shows that, for PDEs with a known and differentiable residual, directional sketching and PDE-aware loss design provide an effective on-the-fly sensitivity signal, rather than merely shifting offline tangent-solve cost into the training loop.

\section{Related Work}
\label{sec:related}

\textbf{Neural operators and physics-informed operator learning.} Neural operators learn mappings between function spaces and have become a standard surrogate-modeling tool for parametric PDEs. Representative architectures include DeepONet~\cite{lu2021learning}, the Fourier neural operator (FNO)~\cite{li2021fourier}, graph neural operators~\cite{li2020neural}, multipole graph neural operators~\cite{li2020multipole}, multiwavelet neural operators~\cite{gupta2021multiwavelet}, and broader neural-operator formulations~\cite{kovachki2023neural}. These methods are typically trained from paired input--solution data and aim to learn the solution map itself. A related line of work incorporates PDE residuals into operator learning, leading to physics-informed DeepONets and physics-informed neural operators~\cite{wang2021learning,li2024physics,karniadakis2021physics}. These approaches reduce dependence on labeled solution data by enforcing the governing PDE at sampled space-time locations. In contrast, our work focuses on derivative-informed training: the residual is imposed not on the predicted solution value, but on Jacobian--vector products of the learned solution operator through the forward sensitivity equation.

\textbf{Sobolev and derivative-informed training.}
Sobolev training augments standard value-based regression with derivative supervision and has been shown to improve sample efficiency and generalization when target derivatives are available~\cite{czarnecki2017sobolev}. In scientific machine learning, derivative information is particularly important because learned surrogates are often embedded in optimization, inverse problems, control, and uncertainty quantification pipelines, where surrogate gradients directly influence downstream decisions. Recent studies have also demonstrated the benefits of Sobolev training in learning elliptic equations~\cite{lu2022sobolev}. More recently, neural-operator methods have adapted Sobolev training to PDE solution maps by supervising Jacobian--vector products of the true solution operator. DINO~\cite{o2024derivative}, DIFNO~\cite{yao2025difno}, and sensitivity-constrained FNO~\cite{behroozi2025sensitivity} compute reference sensitivities by solving forward sensitivity equations with a classical PDE solver and then train the neural operator to match these offline-generated tangent labels. These methods demonstrate the value of derivative supervision, but they share the same offline pipeline: derivative data must be generated and stored before training. Our method preserves the Sobolev-training objective while removing the offline derivative-label generation stage by enforcing the forward sensitivity equation directly during training.

\textbf{Physics-informed residual losses and conditioning.} Our method is also connected to the large literature on physics-informed neural networks (PINNs)~\cite{raissi2019physics,karniadakis2021physics}. PINNs train neural networks by penalizing PDE residuals at randomly sampled collocation points, and physics-informed neural operators extend this principle to operator learning~\cite{wang2021learning,li2024physics}. However, it is now well understood that raw physics-informed residual losses can be difficult to optimize. Prior work has identified gradient pathologies and loss-scale imbalance~\cite{wang2021understanding}, spectral bias and stiffness effects~\cite{wang2022when}, and failure modes for convection-, reaction-, and diffusion-dominated PDEs~\cite{krishnapriyan2021characterizing}. Recent work further interprets these issues through the lens of operator conditioning and preconditioning~\cite{de2024operator} or addresses them with curvature-aware optimizer corrections~\cite{an2026lightweight}. Our analysis indicates that the same conditioning issue arises at the derivative level: a raw forward-sensitivity residual may be computationally cheap but statistically and numerically ineffective unless the tangent operator is properly conditioned. This motivates our operator-aware loss-design rules for different PDEs.

\textbf{Randomized sketching and Jacobian-vector products.} The computational feasibility of sTCL relies on directional sketching. Instead of materializing the full surrogate Jacobian, we sample a small number of input perturbation directions and evaluate the corresponding Jacobian--vector products using forward-mode automatic differentiation. This connects our method to randomized numerical linear algebra and sketching methods, where low-dimensional random projections are used to estimate high-dimensional linear or quadratic quantities efficiently~\cite{hutchinson1989stochastic,halko2011finding,woodruff2014sketching,martinsson2020randomized}. It is also closely related to the use of Jacobian--vector and vector--Jacobian products in automatic differentiation and large-scale optimization~\cite{pearlmutter1994fast,griewank2008evaluating}. In our setting, sketching has a specific physics-informed interpretation: rather than sampling collocation points as in PINNs, sTCL samples directions in the input function space and enforces the forward sensitivity equation in expectation.

\textbf{Position of this work.} Existing derivative-informed neural-operator methods show that Jacobian supervision improves learned PDE surrogates, while physics-informed methods show that governing equations can be enforced through residual losses. This paper combines these two ideas at the derivative level. The resulting method is derivative-label-free, operates on the fly during neural-operator training, and is computationally practical because the derivative penalty is sketched. The main additional insight is that sketching alone is not enough: conditioning of the forward sensitivity equation is the key factor that determines whether the on-the-fly derivative signal is useful.

\section{Methodology}
\label{sec:method}

We now present the proposed sketched tangent consistency loss (sTCL). The method is built on a simple observation: offline derivative-informed neural-operator methods and our on-the-fly method use the same mathematical object, namely the forward sensitivity equation. The difference is how this equation is used. Offline Sobolev methods solve it with a classical PDE solver to generate derivative labels, whereas sTCL enforces it directly as a stochastic consistency constraint during training.

\subsection{Assumptions}
\label{sec:method_assumptions}

sTCL assumes that (i) the governing PDE of the forward problem is known; (ii) its residual can be evaluated differentiably at the surrogate prediction; and (iii) the linearization of the residual, i.e., its tangent actions with respect to the state and the input, is accessible, either explicitly or by automatic differentiation, without forming full tangent matrices. Thus sTCL removes the need for offline tangent labels, but not the need for the governing equations; purely black-box simulators and systems with unknown physics are outside the scope of this work.

\subsection{Neural operators and forward sensitivity equation}
\label{sec:method_fse}

Let $\mathcal{X}$ and $\mathcal{Y}$ be Banach spaces of input functions and solution functions, respectively. We consider a parametric PDE written abstractly as
\begin{equation}
 F(u,f)=0,
 \qquad f\in \mathcal{X},\quad u\in \mathcal{Y},
    \label{eq:pde_abstract}
\end{equation}
where $F:\mathcal{Y}\times \mathcal{X}\to \mathcal{Z}$ maps the state and input to a residual space $\mathcal{Z}$. When \eqref{eq:pde_abstract} is well posed, it defines a solution operator $S:\mathcal{X}\to \mathcal{Y}$, $u=S(f)$, and a neural operator $S_\theta:\mathcal{X}\to \mathcal{Y}$ is trained to approximate $S$ from paired input--solution data $\{(f_i,u_i)\}_{i=1}^{n}$~\cite{lu2021learning,li2021fourier,li2020neural,li2020multipole,gupta2021multiwavelet,kovachki2023neural}.

Assume that $F$ is Fréchet differentiable with respect to both arguments. For a perturbation direction $v\in \mathcal{X}$, define the directional sensitivity $w = DS(f)[v] \in \mathcal{Y}$, where $DS(f):\mathcal{X}\to \mathcal{Y}$ denotes the Fréchet derivative of the solution operator. Differentiating $F(S(f),f)=0$ in the direction $v$ gives the forward sensitivity equation
\begin{equation}
 D_uF(u,f)[w] + D_fF(u,f)[v] = 0,
 \qquad u=S(f),\quad w=DS(f)[v].
    \label{eq:forward_sensitivity}
\end{equation}
For compactness, we write $\mathcal{A}(u,f) := D_uF(u,f)$ and  $\mathcal{B}(u,f) := D_fF(u,f)$, so that \eqref{eq:forward_sensitivity} becomes
\begin{equation*}
 \mathcal{A}(u,f)w + \mathcal{B}(u,f)v = 0.
\end{equation*}
Here $\mathcal{A}(u,f):\mathcal{Y}\to \mathcal{Z}$ is the tangent operator with respect to the state, and $\mathcal{B}(u,f):\mathcal{X}\to \mathcal{Z}$ describes how the PDE residual changes with respect to the input. In a discretized PDE, $\mathcal{A}$ and $\mathcal{B}$ become the usual Jacobian matrices of the residual with respect to the state and input variables.

The role of derivative-informed training is to improve the learned Jacobian $DS_\theta(f)$, not only the solution value $S_\theta(f)$. This matters whenever the neural operator is used as a differentiable surrogate. For example, for a reduced objective $\Phi(f)=J(S(f),f)$, the surrogate gradient contains the adjoint action of the learned Jacobian,
\begin{equation*}
 \nabla_f \Phi_\theta(f)
 = DS_\theta(f)^\ast \nabla_u J(S_\theta(f),f)
 +
 \nabla_f J(S_\theta(f),f),
\end{equation*}
where $DS_\theta(f)^\ast$ denotes the adjoint of the surrogate derivative. Therefore, accurate solution prediction alone does not guarantee accurate downstream gradients.

\subsection{From offline Sobolev supervision to on-the-fly sketched tangent consistency}
\label{sec:method_stcl}

Sobolev training augments value-based supervision with derivative supervision. In neural-operator training, existing offline derivative-informed methods obtain this supervision by solving the forward sensitivity equation \eqref{eq:forward_sensitivity} for selected inputs and perturbation directions before training, then storing the resulting tangent labels. This can provide accurate derivative targets, but it turns sensitivity information into a precomputed dataset: changing the training distribution, resolution, dataset size, or sketch basis requires regenerating and storing new tangent solutions.

sTCL removes this offline tangent-label stage. Instead of asking a PDE solver to produce $DS(f)[v]$ in advance, we differentiate the current neural operator in randomly sampled directions and penalize violations of the forward sensitivity equation. The population sketched tangent consistency loss is
\begin{equation*}
 \mathcal{L}_{\mathrm{sTCL}}(\theta)
 =
 \mathbb{E}_{f\sim \rho}
 \mathbb{E}_{v\sim \nu}
 \left[
 \left\| r_\theta(f,v)
 \right\|_{\mathcal{Z}}^2
 \right],
\end{equation*}
where $\rho$ is the input distribution, $\nu$ is a sketch distribution over perturbation directions, and $r_\theta(f,v)$ is the tangent residual. To define it, let $u_\theta=S_\theta(f)$ be the surrogate prediction and $w_\theta(f,v)=DS_\theta(f)[v]$ be the surrogate directional derivative. Then
\begin{equation*}
 r_\theta(f,v)
 =
 \mathcal{A}(u_\theta,f)w_\theta(f,v)
 +
 \mathcal{B}(u_\theta,f)v.
\end{equation*}
If $S_\theta=S$, then $u_\theta=S(f)$, $w_\theta=DS(f)[v]$, and $r_\theta(f,v)=0$ for every direction $v$. Thus, the forward sensitivity equation provides a derivative-label-free consistency condition for the learned Jacobian.

In practice, for a minibatch $\mathcal{I}$ and $q$ independently sampled directions $\{v_{i,k}\}_{k=1}^{q}$ for each $f_i\in \mathcal{I}$, we use the estimator
\begin{equation}
 \widehat{\mathcal{L}}_{\mathrm{sTCL}}(\theta;\mathcal{I})
 =
 \frac{1}{|\mathcal{I}|q}
 \sum_{f_i\in \mathcal{I}}
 \sum_{k=1}^{q}
 \left\|
 \mathcal{A}(S_\theta(f_i),f_i)DS_\theta(f_i)[v_{i,k}]
 +
 \mathcal{B}(S_\theta(f_i),f_i)v_{i,k}
 \right\|_{\mathcal{Z}}^2.
    \label{eq:empirical_stcl}
\end{equation}
The training objective at update $t$ is
\begin{equation*}
 \widehat{\mathcal{L}}_{\mathrm{total}}(\theta;\mathcal{I})
 =
 \widehat{\mathcal{L}}_{\mathrm{data}}(\theta;\mathcal{I})
 +
 \gamma_t
 \widehat{\mathcal{L}}_{\mathrm{sTCL}}(\theta;\mathcal{I}),
\end{equation*}
where $\widehat{\mathcal{L}}_{\mathrm{data}}$ is the standard minibatch data-fitting loss. All derivative-informed experiments use the same EMA normalization $\gamma_t=\lambda\overline{\mathcal{L}}_{\mathrm{data},t}/\max(\overline{\mathcal{L}}_{\mathrm{deriv},t},\varepsilon)$, where the bars denote exponential moving averages with decay $0.99$, $\varepsilon>0$ is a small numerical stabilizer, and $\mathcal{L}_{\mathrm{deriv}}$ denotes the sTCL loss here or the JVP-matching loss for offline DI. Thus $\lambda>0$ specifies the target derivative-to-data loss ratio, while the actual multiplier $\gamma_t$ varies during training. The Jacobian--vector product $DS_\theta(f_i)[v_{i,k}]$ is computed by forward-mode automatic differentiation, so the full Jacobian $DS_\theta(f_i)$ is never materialized.

The sketching interpretation becomes explicit after discretization. Let $R_\theta(f)$ denote the linear map from an input perturbation $v$ to the tangent residual,
\begin{equation*}
 R_\theta(f)v
 = A_\theta(f)J_\theta(f)v
 + B_\theta(f)v,
\end{equation*}
where $J_\theta(f)$ is the discrete Jacobian of $S_\theta$, $A_\theta(f)$ is the discretized state tangent operator evaluated at $S_\theta(f)$, and $B_\theta(f)$ is the discretized input tangent operator.
\begin{proposition}[Directional sketching estimates residual energy]
If the sketch direction $v$ is isotropic, i.e.\ $\mathbb{E}[vv^\top]=I$, then
\[
 \mathbb{E}_{v}\|R_\theta(f)v\|_2^2
 =
 \|R_\theta(f)\|_{\mathrm{F}}^2 .
\]
\end{proposition}
This is the standard Hutchinson trace identity applied to $R_\theta(f)^\top R_\theta(f)$~\cite{hutchinson1989stochastic,halko2011finding,woodruff2014sketching,martinsson2020randomized}. Thus, sTCL enforces the forward sensitivity equation in expectation over random input directions. This is the derivative-level analogue of physics-informed training: instead of randomly sampling space-time collocation points and enforcing the PDE residual on average, sTCL randomly samples perturbation directions in the input space and enforces the forward sensitivity equation on average.

\begin{remark}[Why sketching makes stochastic training practical]
The estimator \eqref{eq:empirical_stcl} is compatible with stochastic neural-network training because its cost scales with the number of sampled directions, not with the input dimension. Each sampled direction requires one Jacobian--vector product through the neural operator and one application of the known tangent operators $\mathcal{A}$ and $\mathcal{B}$. Hence, for $q$ sketches,
\begin{equation*}
 \mathrm{Cost}(\widehat{\mathcal{L}}_{\mathrm{sTCL}})
 \approx q\,\mathrm{Cost}(\mathrm{JVP}_{S_\theta})
 + q\,\mathrm{Cost}(\mathcal{A},\mathcal{B})
 +
 \mathrm{Cost}(\mathrm{backward}).
\end{equation*}
For FNO-type architectures, a forward-mode Jacobian--vector product is comparable to a small constant multiple of a forward pass, while the tangent-operator applications are typically FFT-, DST-, or stencil-based operations~\cite{trefethen2000spectral,leveque2007finite}. Therefore, sTCL does not require full-Jacobian materialization, does not require offline derivative labels, and keeps the derivative penalty at the same computational order as the data-fitting loss up to the sketch factor $q$.
\end{remark}

\subsection{Conditioning and operator-aware loss design}
\label{sec:method_conditioning}

Sketching makes the on-the-fly derivative loss computationally feasible, but it does not guarantee that the loss is well conditioned. The conditioning of sTCL is governed by the tangent operator in the forward sensitivity equation. To see this, fix $(f,v)$ and locally freeze the surrogate state $u_\theta=S_\theta(f)$. Write $A=\mathcal{A}(u_\theta,f)$, $b=\mathcal{B}(u_\theta,f)v$, and $w=DS_\theta(f)[v]$. The raw tangent residual loss is $\ell_{\mathrm{raw}}(w)=\|Aw+b\|_{\mathcal{Z}}^2$; its Hessian with respect to $w$ is $2A^\ast A$ in Hilbert-space notation, and $2A^\top A$ after discretization. Therefore, when $A$ is nonsingular, the condition number is
\begin{equation}
 \kappa(\nabla_w^2 \ell_{\mathrm{raw}})
 =
 \kappa(A^\top A)
 =
 \kappa(A)^2.
    \label{eq:condition_squared}
\end{equation}
This identity indicates that the raw forward-sensitivity residual can be much harder to optimize than the tangent equation itself: the residual loss squares the singular-value spectrum of the tangent operator. If $A$ is stiff, nearly singular, or indefinite with small resonant modes, then the gradient signal from the raw residual can be poorly scaled or misleading, making first-order training methods struggle even though the residual is mathematically correct.

We therefore introduce operator-aware loss design. The residual-form preconditioned loss is
\begin{equation}
 \ell_M(w)
 =
 \langle Aw+b,\,M(Aw+b)\rangle_{\mathcal{Z}},
 \qquad M\succeq 0,
    \label{eq:preconditioned_residual_loss}
\end{equation}
where $M:\mathcal{Z}\to \mathcal{Z}$ is a positive semidefinite preconditioner. Its local Hessian is $2A^\ast M A$ in Hilbert-space notation, or $2A^\top M A$ after discretization, so the design goal is to choose a cheap $M$ such that $\kappa(A^\top M A)\ll \kappa(A^\top A)$. For symmetric positive-definite elliptic tangent operators, a useful residual preconditioner can behave like an inverse of $A$, but computing $A^{-1}$ exactly would amount to solving the tangent equation. Instead, we use cheap spectrally equivalent approximations, such as DST-based inverse-Laplacian or shifted-Laplacian preconditioners, that reduce the condition number of the residual loss without solving the forward sensitivity equation.

For indefinite or nearly singular tangent operators, a simple positive residual preconditioner may still be insufficient. If $A$ has near-null or resonant modes, then $A^\top M A$ can provide weak curvature along physically important directions, and minimizing the raw residual may suppress components of the sensitivity. For the Helmholtz experiments, we therefore use a short-iteration MINRES target-matching loss. Let $M_0$ denote a cheap shifted-Laplacian base preconditioner, and let $P_m(A,M_0)$ denote the approximate inverse action obtained by applying $m$ steps of preconditioned MINRES. Writing the tangent equation as $Aw=g$ with $g=-b$, we form an on-the-fly approximate target
\begin{equation}
 \hat w_m = P_m(A,M_0)g
 \qquad\text{and minimize}\qquad
 \ell_{\mathrm{MINRES\text{-}target}}(w)=\|w-\hat w_m\|_{\mathcal{Y}}^2 .
    \label{eq:minres_target_matching}
\end{equation}
Here $m$ is fixed and small, typically $m\leq 5$, so $P_m(A,M_0)$ should be viewed as a cheap on-the-fly target constructor rather than an accurate tangent solver. If $P_m(A,M_0)\approx A^{-1}$, then \eqref{eq:minres_target_matching} behaves like a well-conditioned sensitivity loss. As $m\to\infty$, the target approaches the exact tangent solution $A^{-1}g$, but sTCL deliberately operates in the small-$m$ regime, where the goal is to improve the conditioning and informativeness of the derivative loss rather than to solve the forward sensitivity equation accurately.

The resulting operator-aware rule is summarized in \Cref{tab:benchmark_conditioning}. The table should be read as a loss-design guideline: use a residual loss when a cheap positive preconditioner can make $A^\top M A$ well conditioned, and use short-iteration MINRES target matching for symmetric indefinite operators with near-resonant modes, as in Helmholtz.

\begin{remark}[Why sTCL does not merely amortize offline tangent-solve cost]
A natural concern is that on-the-fly derivative-informed training may simply move the offline tangent-solve cost into the training loop. sTCL avoids this. Offline Sobolev methods solve the forward sensitivity equation to generate accurate tangent labels before training. In contrast, sTCL uses on-the-fly consistency and conditioning to construct a stochastic derivative-informed signal. The residual preconditioners in \eqref{eq:preconditioned_residual_loss} are cheap operator-aware transformations designed to reduce the condition number of $A^\top M A$, not to compute $A^{-1}b$. Likewise, the MINRES target constructor in \eqref{eq:minres_target_matching} uses a fixed small number of iterations to match the accuracy level needed during training, rather than a high-accuracy tangent solve aimed at solver-level accurate sensitivities. Thus, sTCL replaces accurate offline tangent solves with sketched on-the-fly consistency and inexpensive conditioning mechanisms, reducing the total computational burden rather than merely amortizing the offline data-generation cost.
\end{remark}

\section{Experiments}
\label{sec:experiments}

We evaluate sTCL against offline DI~\cite{yao2025difno}\footnote{Throughout, ``offline DI'' denotes our offline derivative-informed comparison baseline (an FNO trained on precomputed forward-sensitivity labels, following~\cite{yao2025difno}). Our offline-DI implementation uses the same PDE-specific probe-direction distribution and the same per-update direction count $q=4$ as sTCL; offline DI samples from a fixed bank of precomputed direction--label pairs, whereas sTCL uses unlabeled directions and constructs its derivative loss on the fly. Thus the comparison uses matched probe construction rather than the original KLE-input/POD-output reduced basis of~\cite{yao2025difno}.} on five PDE benchmarks spanning the operator classes in \cref{tab:benchmark_conditioning}. We use offline DI as the representative offline derivative-informed baseline because it shares the same FNO backbone and the same central supervision mechanism as DINO~\cite{o2024derivative} and sensitivity-constrained FNO~\cite{behroozi2025sensitivity}: offline solution of forward sensitivity equations followed by stored Jacobian--vector labels. Thus the comparison isolates the main distinction studied here---offline tangent-label supervision versus on-the-fly sketched tangent consistency---rather than differences in architecture or training infrastructure. Pure FNO is therefore the shared backbone and the no-derivative-supervision ablation, not the principal competing baseline; we adopt the DIFNO backbone for sTCL only to keep this comparison controlled, since sTCL does not depend on the architecture. \Cref{tab:benchmark_conditioning} gives the benchmark equations, tangent classes, tangent features, and sTCL loss forms used in the main comparison; full discretization, data-generation, optimizer, and training details are deferred to \cref{app:experiment_details,app:per_cell_configs}. In brief, all main comparisons use FNO backbones, Adam lr$=10^{-3}$, batch size $32$, sketch count $q=4$, three training seeds, and matched PDE-specific direction sampling between sTCL and offline DI. Helmholtz, nonlinear diffusion, and Burgers are trained for $2000$ epochs; Allen--Cahn is trained for $1500$ epochs; and Navier--Stokes is trained for $5000$ epochs. Within each PDE, all methods share the same backbone, grid, and training protocol, and both derivative-informed methods use the same EMA-normalized interpretation of $\lambda$. The weight $\lambda$ is selected on validation grids detailed in \cref{app:per_cell_configs}.

\begin{table}[t]
\centering
\scriptsize
\caption{Benchmarks, tangent-operator classes, and sTCL forms used in the main experiments. The table combines the loss-design rule with the experimental suite; boundary conditions, data distributions, solvers, tangent equations, and implementation details are in \cref{app:experiment_details}. In the equation/map column, blue marks the operator input and red marks the response.}
\label{tab:benchmark_conditioning}
\setlength{\tabcolsep}{2pt}
\begin{tabular}{p{0.13\linewidth} p{0.30\linewidth} p{0.16\linewidth} p{0.21\linewidth} p{0.15\linewidth}}
\toprule
Benchmark & Equation / map & Tangent class & Tangent feature & sTCL form \\
\midrule
Nonlinear diffusion &
$-\nabla\!\cdot(e^{\textcolor{blue}{a}}\nabla \textcolor{red}{u})+\textcolor{red}{u}^3=f$ &
SPD elliptic &
Stiff elliptic spectrum &
DST inverse-Laplacian residual \\
Helmholtz &
$(-\Delta-\kappa^2e^{2\textcolor{blue}{a}})\textcolor{red}{u}=f$ &
Indefinite elliptic &
    Near-resonant modes &
    MINRES target matching \\
Burgers &
$\textcolor{red}{u}_t+\textcolor{red}{u}\,\textcolor{red}{u}_x=\nu \textcolor{red}{u}_{xx}+\textcolor{blue}{f(t)}$ &
Parabolic, non-self-adjoint &
Advective dynamics &
RHS-normalized space--time residual \\
Allen--Cahn &
$\begin{gathered}
\textcolor{red}{u}_t=\varepsilon^2\textcolor{red}{u}_{xx}+\textcolor{red}{u}-\textcolor{red}{u}^3,\\
\textcolor{red}{u}(x,0)=\textcolor{blue}{a(x)}
\end{gathered}$ &
Parabolic, phase-field &
Interface-sensitive coarsening &
Raw space--time $L^2$ residual \\
Navier--Stokes &
$\begin{gathered}
-\mu\Delta\textcolor{red}{y}+(\textcolor{red}{y}\!\cdot\!\nabla)\textcolor{red}{y}+\nabla\textcolor{red}{p}=\textcolor{blue}{u},\\
\nabla\!\cdot\!\textcolor{red}{y}=0
\end{gathered}$ &
Non-self-adjoint saddle point &
Pressure coupling and advection &
Leray--Oseen-preconditioned residual \\
\bottomrule
\end{tabular}
\end{table}

\subsection{On-the-fly sTCL versus offline DI on function values}

\paragraph{Function-value accuracy (\cref{tab:master_5pde}).}
\Cref{tab:master_5pde} reports the mean per-sample relative response error
\[
\bar E_z \;=\; \frac{1}{N_{\mathrm{test}}}\sum_{i=1}^{N_{\mathrm{test}}}
\frac{\|z_i - S_\theta(a_i)\|_2}{\|z_i\|_2},
\]
where $z_i=S(a_i)$ is the reference response for the held-out input $a_i$ and $S_\theta(a_i)$ is the neural-operator prediction. We report $100\bar E_z$ for the best-epoch checkpoint, i.e., the checkpoint with the lowest validation error, and average the resulting test errors over three seeds. For Navier--Stokes, $a_i$ is the distributed forcing and $z_i$ is the two-component velocity; the pressure is mean-gauged, included as an auxiliary training output, and excluded from this table.

\Cref{tab:master_5pde} shows that incorporating derivative information improves the FNO surrogate in most regimes. More importantly, under the same training protocol, sTCL attains accuracy close to that of offline DI without requiring any offline tangent labels: its relative error is within $1.2{\times}$ that of offline DI in $11$ of the $15$ cells and within $1.5{\times}$ in $14$ of the $15$ cells.

The remaining differences are PDE- and data-regime-dependent rather than evidence of a uniformly dominant method. Offline DI has lower mean errors in all three regimes for nonlinear diffusion and Navier--Stokes. On Helmholtz, offline DI is lower at $N{=}128$, whereas sTCL is lower at $N{=}512$ and $N{=}1024$; on Burgers, offline DI is lower at $N{=}128$, whereas sTCL is lower at $N{=}32$ and $N{=}512$; and on Allen--Cahn, offline DI is lower at $N{=}1024$, whereas sTCL is lower at $N{=}512$ and $N{=}2048$. The two methods are close on Helmholtz at $N{=}128$ and Burgers at $N{=}128$ and $N{=}512$, and remain on the same sub-percent accuracy scale across all Allen--Cahn and Navier--Stokes regimes. Moreover, sTCL reduces the pure-FNO mean function-value error in every one of the $15$ comparison cells. Thus the function-value results support the central claim that on-the-fly sketched tangent consistency can replace stored offline tangent labels at comparable accuracy, while eliminating the offline tangent-data pipeline.

\begin{table}[t]
\centering
\small
\caption{Function-value test relative error (\%), reported as mean $\pm$ standard deviation over three seeds. All cells use Adam with lr$=10^{-3}$ and batch size $B=32$; derivative-informed cells use sketch count $q=4$. Helmholtz, nonlinear diffusion, and Burgers are trained for $2000$ epochs; Allen--Cahn is trained for $1500$ epochs; and Navier--Stokes is trained for $5000$ epochs. All derivative-informed cells use the same EMA-normalized weighting. The selected $\lambda$ values are listed in \cref{tab:per_cell_configs}. The Navier--Stokes rows report velocity error. Within each PDE, sTCL and offline DI use matched probe-direction distributions and the same per-update direction count, so the comparison does not confound the supervision mode with different probe constructions.}
\label{tab:master_5pde}
\setlength{\tabcolsep}{4pt}
\begin{tabular}{l c r r r}
\toprule
PDE & $N$ & Pure FNO & sTCL & Offline DI \\
\midrule
\multirow{3}{*}{Helmholtz}                & 128 & $\phantom{0}7.95\pm0.54$ & $\phantom{0}7.48\pm0.54$ & $\phantom{0}7.13\pm0.90$ \\
 & 512 & $\phantom{0}3.92\pm0.46$ & $\phantom{0}2.25\pm0.31$ & $\phantom{0}3.08\pm1.37$ \\
 & 1024 & $\phantom{0}2.09\pm0.25$ & $\phantom{0}1.51\pm0.17$ & $\phantom{0}1.83\pm0.75$ \\
\midrule
\multirow{3}{*}{Nonlin.\ diffusion}       & 128 & $\phantom{0}9.71\pm0.15$ & $\phantom{0}4.77\pm0.17$ & $\phantom{0}3.43\pm0.16$ \\
 & 512 & $\phantom{0}4.49\pm0.16$ & $\phantom{0}2.26\pm0.07$ & $\phantom{0}1.52\pm0.03$ \\
 & 1024 & $\phantom{0}2.98\pm0.05$ & $\phantom{0}1.65\pm0.03$ & $\phantom{0}1.07\pm0.02$ \\
\midrule
\multirow{3}{*}{Burgers}                  & 32  & $17.75\pm3.10$ & $\phantom{0}7.65\pm0.70$ & $\phantom{0}8.14\pm0.48$ \\
 & 128 & $\phantom{0}3.01\pm0.24$ & $\phantom{0}1.72\pm0.07$ & $\phantom{0}1.69\pm0.05$ \\
 & 512 & $\phantom{0}0.58\pm0.06$ & $\phantom{0}0.52\pm0.09$ & $\phantom{0}0.57\pm0.03$ \\
\midrule
\multirow{3}{*}{Allen--Cahn}              & 512  & $\phantom{0}0.94\pm0.02$ & $\phantom{0}0.59\pm0.06$ & $\phantom{0}0.60\pm0.04$ \\
 & 1024 & $\phantom{0}0.50\pm0.01$ & $\phantom{0}0.40\pm0.03$ & $\phantom{0}0.35\pm0.01$ \\
 & 2048 & $\phantom{0}0.27\pm0.01$ & $\phantom{0}0.25\pm0.03$ & $\phantom{0}0.27\pm0.01$ \\
\midrule
\multirow{3}{*}{Navier--Stokes}        & 512  & $\phantom{0}0.619\pm0.009$ & $\phantom{0}0.400\pm0.029$ & $\phantom{0}0.324\pm0.013$ \\
 & 1024 & $\phantom{0}0.392\pm0.003$ & $\phantom{0}0.283\pm0.016$ & $\phantom{0}0.236\pm0.009$ \\
 & 2048 & $\phantom{0}0.264\pm0.012$ & $\phantom{0}0.198\pm0.042$ & $\phantom{0}0.172\pm0.005$ \\
\bottomrule
\end{tabular}
\end{table}

\paragraph{Training time and offline data generation.}
sTCL and offline DI have per-job training times of the same order (\cref{tab:wallclock_per_pde}), and the offline-DI times exclude the separate cost of generating and storing tangent labels (\cref{tab:offline_cost} in \cref{app:offline_cost}), which sTCL avoids entirely.

\begin{table}[t]
\centering
\small
\caption{Representative full-budget per-training-job wall-clock at $N{=}512$ on a single RTX~PRO~6000 Blackwell GPU; times are not averaged over training seeds. Helmholtz, nonlinear diffusion, and Burgers use $2000$ epochs; Allen--Cahn uses $1500$ epochs; and Navier--Stokes uses $5000$ epochs. ``offline-DI training'' excludes offline CPU tangent-data generation.}
\label{tab:wallclock_per_pde}
\begin{tabular}{l r r r}
\toprule
PDE & Pure FNO & sTCL (on the fly) & Offline DI (training only) \\
\midrule
Helmholtz & $192$\,s & $2412$\,s (MINRES-5) & $2036$\,s \\
Nonlin.\ diffusion & $192$\,s & $1976$\,s (DST-precond.) & $1796$\,s \\
Burgers & $408$\,s & $2144$\,s (normalized residual) & $2208$\,s \\
Allen--Cahn & $291$\,s & $2295$\,s (raw $L^2$) & $1911$\,s \\
Navier--Stokes & $582$\,s & $9990$\,s (Leray--Oseen) & $4810$\,s \\
\bottomrule
\end{tabular}
\end{table}

\subsection{On-the-fly sTCL versus offline DI on Jacobians}

\paragraph{Jacobian relative error (\cref{tab:jac_err}).}
The main motivation for derivative-informed training is sensitivity accuracy, not only function-value accuracy. We therefore evaluate the learned Jacobian using a per-(sample, direction) relative error
\[
\bar E \;=\; \frac{1}{N_{\mathrm{test}}\,n_{\mathrm{dirs}}}\sum_{i=1}^{N_{\mathrm{test}}}\sum_{d=1}^{n_{\mathrm{dirs}}}
\frac{\|J(a_i)v_d - J_\theta(a_i)v_d\|_2}{\|J(a_i)v_d\|_2},
\]
where $\{a_i\}_{i=1}^{N_{\mathrm{test}}}$ is the held-out test set of $N_{\mathrm{test}}$ input fields, $\{v_d\}_{d=1}^{n_{\mathrm{dirs}}}$ are fixed random probe directions in input space, $J(a_i)v_d$ is the true Jacobian--vector product obtained from the discretized forward sensitivity equation using the same tangent-solver pipeline that generates offline-DI training labels, and $J_\theta(a_i)v_d$ is the surrogate Jacobian--vector product computed by forward-mode autodiff (\texttt{torch.func.jvp}) through the trained operator $S_\theta$ (further details in \cref{app:details}).

The Jacobian results show that sTCL and offline DI lie in a comparable accuracy range, with PDE-dependent differences. On Burgers, offline DI is lower at $N{=}32$ and $N{=}128$, but at $N{=}512$ sTCL is markedly more stable than offline DI: $73.53\pm0.60\%$ versus $178.08\pm15.45\%$. On Helmholtz, offline DI is slightly lower at $N{=}128$, while sTCL is lower at $N{=}512$ and $N{=}1024$; both derivative-informed methods remain below the FNO baseline at all three sizes. On nonlinear diffusion, offline DI gives smaller Jacobian errors, but sTCL remains well below the pure-FNO baseline at every cell. On Allen--Cahn, both derivative-informed methods reduce the pure-FNO Jacobian error by a large margin, and sTCL remains close to offline DI across all three training sizes. On Navier--Stokes, offline DI gives smaller velocity-JVP errors, while sTCL remains below the pure-FNO baseline in all three regimes and on the same sub-percent accuracy scale as offline DI. These results support the interpretation that sTCL provides an effective sensitivity signal on the fly: it does not uniformly beat offline DI on Jacobian error, but it recovers much of offline DI's improvement over pure FNO without constructing or storing an offline tangent dataset.

\begin{table}[t]
\centering
\small
\caption{Jacobian relative error (\%), reported as mean $\pm$ standard deviation over three seeds across all $15$ benchmark cells. The training budgets match \cref{tab:master_5pde}: Helmholtz, nonlinear diffusion, and Burgers are trained for $2000$ epochs, Allen--Cahn for $1500$ epochs, and Navier--Stokes for $5000$ epochs. The reported metric is the per-(sample, direction) average $\bar E$.}
\label{tab:jac_err}
\setlength{\tabcolsep}{4pt}
\begin{tabular}{l c r r r}
\toprule
PDE & $N$ & Pure FNO & sTCL & Offline DI \\
\midrule
\multirow{3}{*}{Helmholtz}                & 128 & $51.64\pm 1.48$ & $47.78\pm 2.41$ & $45.99\pm 4.64$ \\
 & 512 & $26.01\pm 3.04$ & $13.83\pm 2.64$ & $19.52\pm 9.03$ \\
 & 1024 & $12.17\pm 0.92$ & $\phantom{0}9.17\pm 1.51$ & $10.67\pm 3.75$ \\
\midrule
\multirow{3}{*}{Nonlin.\ diffusion}       & 128 & $29.74\pm 0.55$ & $16.06\pm 0.62$ & $11.65\pm 0.50$ \\
 & 512 & $15.87\pm 0.30$ & $\phantom{0}8.38\pm 0.15$ & $\phantom{0}5.36\pm 0.05$ \\
 & 1024 & $10.88\pm 0.17$ & $\phantom{0}6.07\pm 0.05$ & $\phantom{0}3.80\pm 0.12$ \\
\midrule
\multirow{3}{*}{Burgers}                  & 32  & $749.68\pm 80.79$ & $\phantom{0}74.76\pm 0.87$ & $\phantom{0}67.83\pm 0.80$ \\
 & 128 & $296.90\pm 18.72$ & $\phantom{0}73.80\pm 0.50$ & $\phantom{0}70.58\pm 0.22$ \\
 & 512 & $182.02\pm 16.97$ & $\phantom{0}73.53\pm 0.60$ & $178.08\pm 15.45$ \\
\midrule
\multirow{3}{*}{Allen--Cahn}              & 512  & $\phantom{0}15.64\pm 1.22$ & $\phantom{0}3.82\pm 0.33$ & $\phantom{0}3.64\pm 0.32$ \\
 & 1024 & $\phantom{0}11.59\pm 1.58$ & $\phantom{0}2.83\pm 0.30$ & $\phantom{0}2.11\pm 0.09$ \\
 & 2048 & $\phantom{0}10.77\pm 1.47$ & $\phantom{0}1.69\pm 0.17$ & $\phantom{0}1.48\pm 0.13$ \\
\midrule
\multirow{3}{*}{Navier--Stokes}        & 512  & $\phantom{0}1.084\pm0.023$ & $\phantom{0}0.570\pm0.024$ & $\phantom{0}0.447\pm0.013$ \\
 & 1024 & $\phantom{0}0.726\pm0.005$ & $\phantom{0}0.458\pm0.024$ & $\phantom{0}0.278\pm0.008$ \\
 & 2048 & $\phantom{0}0.503\pm0.022$ & $\phantom{0}0.282\pm0.014$ & $\phantom{0}0.249\pm0.007$ \\
\bottomrule
\end{tabular}
\end{table}

\subsection{Conditioning strategies}

The main comparisons use a conditioning strategy matched to each tangent operator, as summarized in \cref{tab:benchmark_conditioning}. Burgers and Allen--Cahn use space--time residuals without an additional spectral preconditioner, whereas nonlinear diffusion, Helmholtz, and Navier--Stokes use a DST inverse-Laplacian metric, MINRES target construction, and Leray--Oseen weighting, respectively. Detailed loss definitions are provided in \cref{app:experiment_details}. \Cref{app:minres_iters} further varies the number of MINRES inner iterations used to build the Helmholtz target and shows that $m=5$ matches $m=25$ within seed variability at substantially lower training cost.

\section{Conclusion}
\label{sec:conclusion}

We introduced \emph{sketched tangent consistency loss} (sTCL), an on-the-fly derivative-informed training objective for neural operators. At each training step, sTCL samples a small number of perturbation directions in the input function space, computes the corresponding surrogate Jacobian--vector products by forward-mode automatic differentiation, and applies a PDE-specific consistency loss for the forward sensitivity equation, so that the derivative supervision comes directly from the governing PDE rather than from precomputed tangent solutions or stored derivative labels. Our study identifies two ingredients for on-the-fly derivative-informed training: \emph{sketching}, which makes the derivative penalty compatible with stochastic training, and operator-aware loss \emph{conditioning}, which makes the resulting sensitivity signal effective across tangent-operator classes. Across five PDE benchmarks spanning parabolic/advective, phase-field parabolic, SPD elliptic, symmetric indefinite, and coupled incompressible saddle-point regimes, sTCL achieves solution and Jacobian accuracy comparable to offline derivative-informed training under the same backbone while eliminating the offline derivative-data pipeline and its associated regeneration cost as the dataset, resolution, or sketch basis changes. These results position sTCL as a derivative-label-free and architecture-agnostic alternative to offline derivative supervision.

Several limitations remain, and we highlight four future directions. (1)~\emph{Known physics.} sTCL requires a known PDE with a differentiable residual and access to its linearization or tangent actions; purely black-box simulators and unknown-physics settings are outside the current scope. (2)~\emph{Hyperparameter search.} We use small validation grids: PDE-level grids for Burgers and Allen--Cahn, and per-$(N,\mathrm{method})$ grids for Helmholtz, nonlinear diffusion, and Navier--Stokes. Broader joint tuning of the learning rate and $\lambda$ may improve both sTCL and offline DI, so the reported results should not be viewed as final performance limits. (3)~\emph{Conditioning strategies.} Stronger operator-aware preconditioners and adaptive inner-iteration schedules for indefinite and coupled systems remain open directions. (4)~\emph{Coupled-system cost.} In Navier--Stokes, sTCL differentiates and evaluates the momentum, incompressibility, and boundary residuals; these additional tangent terms increase the per-update cost relative to PDEs with fewer residual components.

\begin{ack}
This work was supported in part by the U.S.\ National Science Foundation under Award No.~2608818. We gratefully acknowledge the NVIDIA Academic Grant Program for providing GPU resources. This work used Jetstream2 at Indiana University through allocation ENG260010 from the Advanced Cyberinfrastructure Coordination Ecosystem: Services \& Support (ACCESS) program, which is supported by U.S.\ National Science Foundation grants \#2138259, \#2138286, \#2138307, \#2137603, and \#2138296. We also acknowledge the National Artificial Intelligence Research Resource (NAIRR) Pilot and Amazon Web Services for computing resources under project NAIRR250089. The authors declare no competing interests.
\end{ack}

\bibliographystyle{unsrtnat}
\bibliography{references}

\appendix
\section{Extended experimental details}
\label{app:details}

\subsection{Experiment protocol and benchmark details}
\label{app:experiment_details}
\label{sec:common_setup}

All surrogate Jacobian--vector products are computed by forward-mode automatic differentiation. Unless otherwise noted, FNO runs use $4$ Fourier layers, width $32$, GELU activations, Adam at learning rate $10^{-3}$, batch size $32$, sketch count $q{=}4$, ReduceLROnPlateau scheduling, gradient clipping at norm $1.0$, and fixed seeds. Helmholtz, nonlinear diffusion, and Burgers are trained for $2000$ epochs in \cref{tab:master_5pde,tab:jac_err}. Allen--Cahn is trained for $1500$ epochs, and Navier--Stokes is trained for $5000$ epochs. The wall-clock timings in \cref{tab:wallclock_per_pde} correspond to these full training budgets. The MINRES ablation in \cref{tab:minres_iters} comes from the $2000$-epoch Helmholtz protocol. The spectral mode count is $8$ for Helmholtz and nonlinear diffusion, and $12$ for Burgers, Allen--Cahn, and Navier--Stokes.

\paragraph{Data splits and test size.}
For each benchmark family, we generate independent training, validation, and test splits with training size $N$ as reported in the tables, $N_{\mathrm{val}}=128$, and $N_{\mathrm{test}}=128$. For Jacobian-error evaluation, the test samples are paired with independently generated probe directions: $8$ directions for Helmholtz, $10$ directions for nonlinear diffusion, $200$ directions for Burgers, $200$ Mat\'ern initial-condition directions for Allen--Cahn, and $16$ solenoidal forcing directions for Navier--Stokes, with Allen--Cahn JVPs evaluated over all stored trajectory frames.

\paragraph{Jacobian evaluation protocol.}
\label{app:jac_eval}
The Jacobian errors in \cref{tab:jac_err} are evaluated post-hoc on the held-out test split. Within each PDE, all three methods are evaluated against the same independently generated test-direction bank, and each entry uses the per-(sample, direction) average $\bar E$ defined alongside the table. Surrogate JVPs are computed by forward-mode automatic differentiation through the trained operator. The reference JVPs follow the same discretized tangent pipeline used to generate the corresponding offline-DI training bank: sparse linearized elliptic solves for Helmholtz and nonlinear diffusion, forward-mode differentiation through the implicit-Picard and Fourier-spectral IMEX solvers for Burgers and Allen--Cahn, respectively, and the stabilized discrete Oseen saddle solve for Navier--Stokes. Allen--Cahn errors include all stored trajectory frames. For Navier--Stokes, the saddle solve includes the pressure tangent, but only the velocity tangent enters the reported metric. The test JVP banks are used only for evaluation. All benchmarks use the best-epoch checkpoint, i.e., the checkpoint with the lowest validation error (full-trajectory error for Allen--Cahn and velocity error for Navier--Stokes); within each benchmark, the same checkpoint supplies the function-value and Jacobian results.

\begin{figure}[t]
\centering
\includegraphics[width=\linewidth]{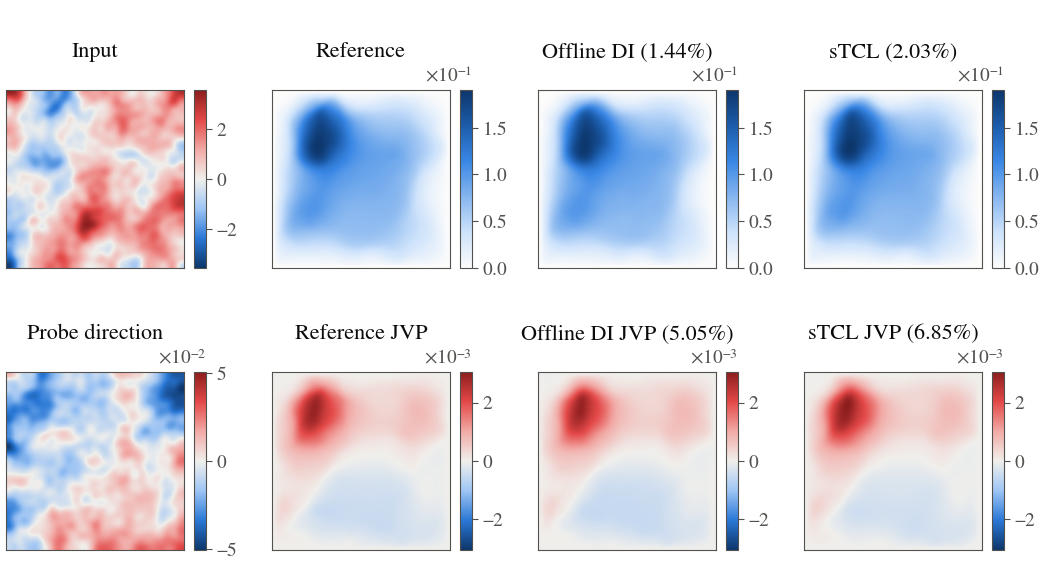}
\caption{Nonlinear diffusion--reaction on a held-out test sample. Top row: input log-diffusivity $a$, reference state $u=S(a)$, and the offline-DI and sTCL predictions $S_\theta(a)$. Bottom row: a test probe direction $\delta a$, the reference JVP $DS(a)[\delta a]$ from the tangent solver, and the surrogate JVPs $DS_\theta(a)[\delta a]$. Prediction panels share the color scale of the corresponding reference panel. Percentages are the relative $L^2$ errors for this sample (top) and this sample--probe pair (bottom), i.e., single terms of the averages reported in \cref{tab:master_5pde,tab:jac_err}.}
\label{fig:overview_nonlinear_diffusion}
\end{figure}

\begin{figure}[t]
\centering
\includegraphics[width=\linewidth]{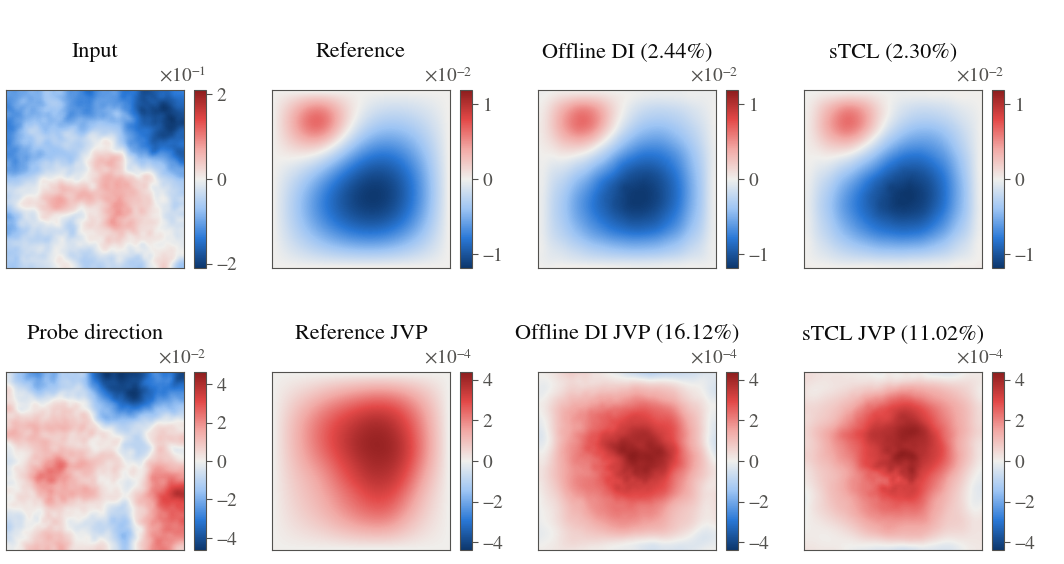}
\caption{Helmholtz on a held-out test sample, with the same layout as \cref{fig:overview_nonlinear_diffusion}. Top row: input coefficient field $a$, reference response $u=S(a)$, and the offline-DI and sTCL predictions. Bottom row: a test probe direction $\delta a$, the reference JVP $DS(a)[\delta a]$, and the surrogate JVPs. Percentages are the relative $L^2$ errors for this sample and sample--probe pair.}
\label{fig:overview_helmholtz}
\end{figure}

\begin{figure}[t]
\centering
\includegraphics[width=\linewidth]{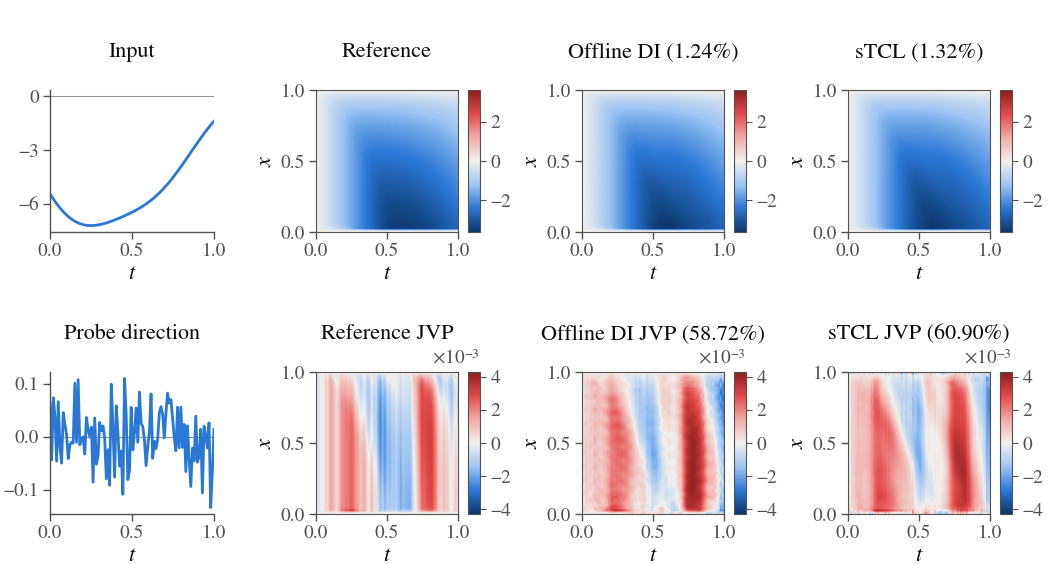}
\caption{Burgers on a held-out test sample, with the same layout as \cref{fig:overview_nonlinear_diffusion}. Top row: the spatially uniform control $f(t)$, the reference trajectory $u(x,t)=S(f)$, and the offline-DI and sTCL predictions; in the space--time panels, $t$ is horizontal and $x$ is vertical. Bottom row: a Gaussian time-grid probe $\delta f(t)$, the reference JVP $DS(f)[\delta f]$, and the surrogate JVPs. Percentages are the relative $L^2$ errors for this sample and sample--probe pair.}
\label{fig:overview_burgers}
\end{figure}

\begin{figure}[t]
\centering
\includegraphics[width=\linewidth]{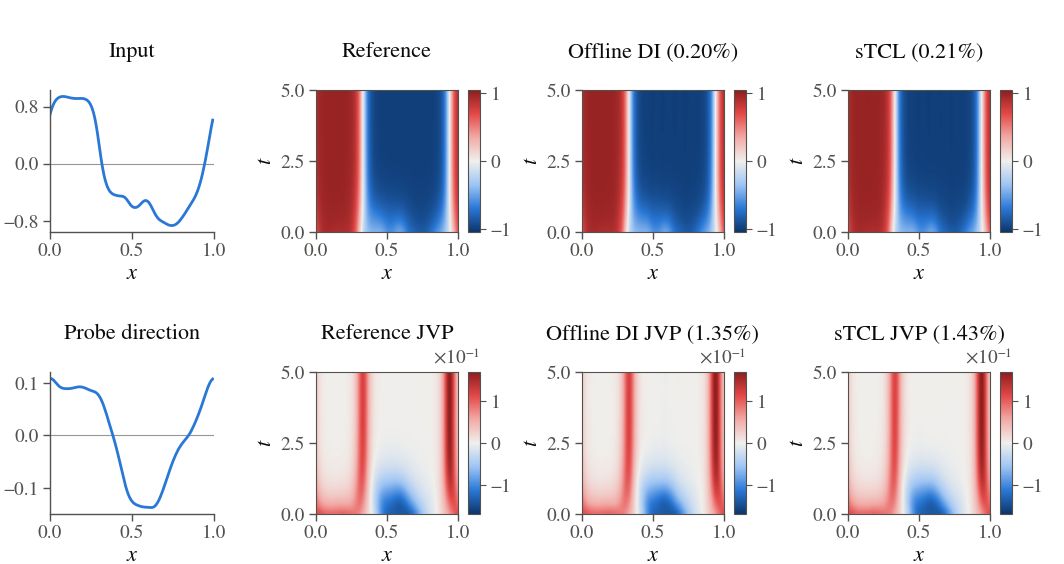}
\caption{Allen--Cahn on a held-out test sample, with the same layout as \cref{fig:overview_nonlinear_diffusion}. Top row: the initial phase field $a(x)$, the reference trajectory $u(x,t)=S(a)$ on $t\in[0,5]$, and the offline-DI and sTCL predictions; in the space--time panels, $x$ is horizontal and $t$ is vertical. Bottom row: a Mat\'ern probe direction $v$, the reference trajectory sensitivity $DS(a)[v]$, and the surrogate JVPs. Percentages are the relative $L^2$ errors for this sample and sample--probe pair.}
\label{fig:overview_allen_cahn}
\end{figure}

\begin{figure}[t]
\centering
\includegraphics[width=\linewidth]{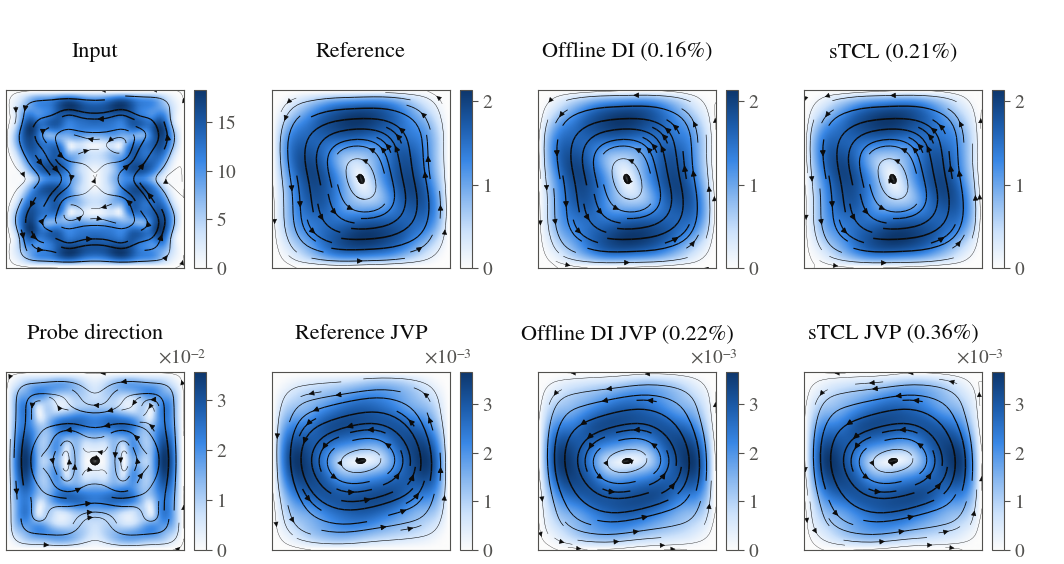}
\caption{Navier--Stokes on a held-out test sample, with the same layout as \cref{fig:overview_nonlinear_diffusion}. Colors show the pointwise magnitude of each vector field and black lines show its streamlines. Top row: the forcing $u$, the reference velocity $y$, and the offline-DI and sTCL velocity predictions. Bottom row: a solenoidal probe direction $s$, the reference velocity JVP $w=Dy(u)[s]$, and the surrogate velocity JVPs. Pressure is not shown. Percentages are the relative $L^2$ velocity errors for this sample and sample--probe pair.}
\label{fig:overview_navier_stokes}
\end{figure}

The benchmark descriptions below use a common organization: setup, tangent equation, conditioning implication, and the sTCL form used in the reported experiments. For each benchmark, \cref{fig:overview_nonlinear_diffusion,fig:overview_helmholtz,fig:overview_burgers,fig:overview_allen_cahn,fig:overview_navier_stokes} show a held-out test input and probe direction together with the reference and predicted solutions and Jacobian--vector products.

\subsubsection{Nonlinear diffusion--reaction.}
\label{sec:nonlinear_diffusion_difno}
\noindent\textbf{Setup.}
The PDE is
\[
 -\nabla\cdot(e^{a(x)}\nabla u(x)) + u(x)^3 = f(x),
 \qquad u|_{\partial\Omega}=0,\quad \Omega=(0,1)^2.
\]
We use a $65\times65$ uniform grid. The centered parameter field is sampled from $C_X=(\tfrac{10}{3}I-\tfrac{1}{30}\Delta)^{-2}$ with Neumann covariance BCs, using a cosine KLE with $40$ modes per coordinate and the $800$ largest-variance mode pairs. The source is the sum of four Gaussian bumps with amplitude $10$, width $0.1$, and centers $(0.25,0.25)$, $(0.75,0.25)$, $(0.25,0.75)$, and $(0.75,0.75)$. Reference states are computed on the $63\times63$ interior grid by a centered finite-difference discretization and Newton iteration.

\noindent\textbf{Tangent equation.}
Differentiating the PDE with respect to an input perturbation $\delta a$ gives
\[
 -\nabla\!\cdot(e^a\nabla \delta u) + 3u^2\delta u
 =
 \nabla\!\cdot(e^a\delta a\nabla u).
\]
Thus the state tangent operator is $A_y(u)\delta u=-\nabla\!\cdot(e^a\nabla\delta u)+3u^2\delta u$. It is symmetric positive definite on $H_0^1(\Omega)$ because $e^a>0$ and $3u^2\ge 0$.

\noindent\textbf{Conditioning.}
The difficulty is not indefiniteness but stiffness. The elliptic operator has eigenvalues that grow like the discrete Laplacian, so the raw residual loss has Hessian $A_y^\top A_y$ and squares the elliptic spectrum as in \cref{eq:condition_squared}. This overweights high-frequency residual components and leaves low-frequency sensitivity components slow to learn. The coefficient $e^a$ also varies over the domain, so a plain Laplacian inverse is only an approximation to a useful inverse-based residual preconditioner.

\noindent\textbf{sTCL form.}
For a unit-normalized probe $\delta a$ drawn from the same $800$-mode cosine KLE as the inputs, we form the discrete tangent residual at the surrogate state $u_\theta(a)$,
\[
r_\theta
=A_y(u_\theta)DS_\theta(a)[\delta a]
-\nabla\!\cdot\!\bigl(e^a\delta a\nabla u_\theta\bigr).
\]
We measure it with
\[
 M = D^{-1/2}(-\Delta_h)^{-1}D^{-1/2},\qquad D=e^a,
\]
where the inverse Dirichlet Laplacian is applied exactly in the DST-I basis. This inexpensive positive metric changes the residual geometry from $A_y^\top A_y$ toward $A_y^\top M A_y$ without solving the tangent equation. At $N{=}1024$, the resulting DST-preconditioned loss lowers the mean per-sample function error from the pure-FNO value $2.98\%$ to $1.65\%$, and the Jacobian error from $10.88\%$ to $6.07\%$.

\subsubsection{Helmholtz.}
\label{sec:helmholtz}
\noindent\textbf{Setup.}
The PDE is
\[
 (-\Delta-\kappa^2 e^{2a(x)})u(x)=f(x),\qquad u|_{\partial\Omega}=0,\quad \kappa=5.
\]
Helmholtz uses the same $65\times65$ Dirichlet grid as nonlinear diffusion. Inputs are sampled from the $40\times40$ cosine expansion $a=\sum_{k,l=0}^{39}\xi_{kl}\,(12.5+0.5\pi^2(k^2+l^2))^{-1}\cos(k\pi x_1)\cos(l\pi x_2)$ with i.i.d.\ $\xi_{kl}\sim\mathcal N(0,1)$, which follows the spectral decay of the centered prior $C_X=(12.5I-0.5\Delta)^{-2}$, and the source is a unit-amplitude Gaussian centered at $(0.258,0.833)$ with width $0.08$. Reference states are obtained by sparse direct solves. To avoid near-resonant response outliers, data generation oversamples the prior and retains draws satisfying $\|u\|_2\leq3$ times the median norm of the candidate pool.

\noindent\textbf{Tangent equation.}
For an input perturbation $\delta a$, the forward sensitivity equation is
\[
 (-\Delta-\kappa^2 e^{2a})\delta u
 = 2\kappa^2 e^{2a}\delta a\,u.
\]
Writing the right-hand side as $g=2\kappa^2 e^{2a}\delta a\,u$, this is $A_y\delta u=g$. The state tangent operator $A_y=-\Delta-\kappa^2 e^{2a}I$ is symmetric and, for almost all sampled inputs, indefinite: for $a\approx0$, $\kappa^2=25$ already lies above the first Dirichlet eigenvalue $2\pi^2$, and spatial variation in $e^{2a}$ perturbs these low modes and can move them toward resonance. The data filter removes the most severe resonant responses, but it does not make the tangent operator positive definite.

\noindent\textbf{Conditioning.}
This case is qualitatively different from SPD diffusion. If a resonant eigenvalue $\lambda_k$ is close to zero, then any residual loss of the form $\|A_y\delta u-g\|^2$ has curvature proportional to $\lambda_k^2$ in that direction. A positive residual preconditioner $M\succeq0$ can rescale modes, but it cannot remove the squared near-zero factor without behaving like an inverse on the resonant subspace. As a result, raw residual minimization can produce a small PDE residual while still missing the physically important sensitivity mode. We therefore match the surrogate JVP to an approximate tangent solution instead of penalizing the residual. Since $A_y$ is symmetric indefinite, this solution is computed with preconditioned MINRES, using the SPD shifted-Laplacian preconditioner $M_0=(-\Delta_h+0.5\kappa^2I)^{-1}$ applied exactly by DST-I.

\noindent\textbf{sTCL form.}
We use MINRES target matching. For each training pair $(a,u)$ in the current mini-batch and each sketched direction $\delta a$, five MINRES iterations preconditioned by $M_0$ construct $\hat w\approx A_y^{-1}g$, and $DS_\theta(a)[\delta a]$ is trained toward $\hat w$. The target is computed only for the current mini-batch and discarded, so no tangent labels are stored. The sweep in \cref{tab:minres_iters} shows that $m{=}5$ is within seed variability of $m{=}10$ and $m{=}25$ on both metrics at substantially lower wall-clock cost.

\subsubsection{Burgers.}
\label{sec:burgers}
\noindent\textbf{Setup.}
The Burgers benchmark is a one-dimensional, time-dependent problem on the spatial interval $x\in(0,1)$, governed by
\[
 u_t + u u_x = \nu u_{xx}+f(t),\qquad \nu=10^{-2},
\]
with $u(0,t)=u(1,t)=0$, $u(x,0)=0$, and final time $T=1$. The spatially uniform control is
\[
f(t)=\sum_{j=1}^{16}c_j\exp\!\left(-\frac{(t-t_j)^2}{2(0.2)^2}\right),
\qquad c_j\sim\mathcal N(0,1.5^2),
\]
where the centers $t_j$ are uniformly spaced on $[0,1]$. The resulting $f(t)$ is broadcast in $x$ as the input channel. Reference trajectories are generated by an implicit finite-difference solver with three Picard iterations per time step, $128$ interior spatial unknowns, and $200$ time steps, then interpolated to the $64\times100$ $(x,t)$ grid used by the FNO. Nonfinite draws and trajectories with $\|u\|_\infty>20$ are rejected and resampled from the same control distribution.

\noindent\textbf{Tangent equation.}
For an input perturbation $\delta f(t)$, the tangent equation is
\[
 w_t + (u w)_x - \nu w_{xx} = \delta f(t),
 \qquad w(0,t)=w(1,t)=0,\quad w(x,0)=0,
\]
so the tangent operator is $A_yw=w_t+(uw)_x-\nu w_{xx}$. It is non-self-adjoint because of the transport term and becomes increasingly advection dominated as $\nu$ decreases.

\noindent\textbf{Conditioning.}
This tangent operator is an initial-value, space--time operator rather than a purely elliptic map. A spatial inverse-Laplacian alone would not account for its time derivative or transport term, so we use no spatial preconditioner ($M=I$). The loss therefore keeps the causal parabolic and advective terms together in the same strong-form residual; integral-form PINN losses are an alternative near shocks~\cite{wang2024coupled}.

\noindent\textbf{sTCL form.}
For each Gaussian time-grid probe $v(t)$, broadcast in space, we compute $w_\theta=DS_\theta(f)[v]$ and
\[
r_\theta=w_{\theta,t}+u_\theta w_{\theta,x}
+w_\theta u_{\theta,x}-\nu w_{\theta,xx}-v.
\]
Centered finite differences are used on the $64\times100$ FNO grid, with a backward time difference at the final frame, and the residual is evaluated only at interior spatial nodes and noninitial frames. The per-sample residual energy is divided by the energy of the corresponding right-hand side; no additional spectral preconditioner or separate tangent boundary/initial block is used. At $N{=}512$, sTCL and offline DI have similar function errors, while their Jacobian errors are $73.53\%$ and $178.08\%$, respectively.

\subsubsection{Allen--Cahn.}
\label{sec:allen_cahn}
\noindent\textbf{Setup.}
The Allen--Cahn benchmark is an unforced one-dimensional phase-field problem on the periodic unit interval,
\[
 u_t=\varepsilon^2 u_{xx}+u-u^3,\qquad x\in(0,1),\qquad u(x,0)=a(x),
\]
\[
 u(0,t)=u(1,t).
\]
We use $\varepsilon=0.03$ and final time $T=5$. The forward map sends the initial phase field $a(x)$ to the full trajectory $u(x,t)$, and all $40$ stored frames, including $t=0$, are used in the data loss and in the evaluation metrics. Initial conditions are sampled from a periodic Mat\'ern Gaussian random field with correlation length $\ell=0.07$ and smoothness exponent $\tau=4$, normalized samplewise and mapped through $\tanh$ into the phase-field range. The reference solver uses $128$ periodic spatial points and Fourier-spectral IMEX Euler stepping, with diffusion implicit in Fourier space and $u-u^3$ explicit.

\noindent\textbf{Tangent equation.}
For an initial-condition perturbation $v=\delta a$ and sensitivity $w=\delta u$, the forward sensitivity equation is
\[
 w_t=\varepsilon^2 w_{xx}+(1-3u^2)w,\qquad x\in(0,1),\qquad w(x,0)=v(x),
\]
\[
 w(0,t)=w(1,t).
\]
Thus the Jacobian of interest maps an initial perturbation to the trajectory sensitivity, $DS(a)[v]=w(\cdot,t)$, over the same stored time frames as the state.

\noindent\textbf{Conditioning.}
The tangent equation is a forward parabolic initial-value problem that couples the time derivative and diffusion with the state-dependent reaction coefficient $1-3u^2$. Its conditioning is therefore governed by the full space--time evolution rather than by a purely spatial elliptic operator. The loss keeps the temporal, diffusive, and reaction terms together and uses the raw space--time $L^2$ tangent residual without additional preconditioning.

\noindent\textbf{sTCL form.}
For unit-normalized Mat\'ern probes with the same $(\ell,\tau)$ as the initial-field sampler, the sTCL loss is the raw space--time $L^2$ tangent residual
\[
 r_\theta(a,v)
 =
 w_{\theta,t}
 -\varepsilon^2 w_{\theta,xx}
 -(1-3u_\theta^2)w_\theta,
 \qquad
 w_\theta=DS_\theta(a)[v],
\]
evaluated over all stored noninitial frames. The neural operator is a two-dimensional FNO over $(t,x)$ with a hard initial-condition ansatz
\[
 u_\theta(a)(t,x)=a(x)+(t/T)N_\theta(a,t/T,x),
\]
so $u_\theta(a)(0,x)=a(x)$ and $D u_\theta(a)[v](0,x)=v(x)$ exactly. In the residual, time derivatives use centered differences at interior stored frames and a second-order backward difference at $T$, while $w_{\theta,xx}$ uses the same Fourier-spectral Laplacian as the data generator. The full-trajectory reference JVPs used for offline DI and Jacobian evaluation are obtained by differentiating through the IMEX solver itself. At $N{=}512$, sTCL lowers the mean field error from $0.94\%$ to $0.59\%$ and the Jacobian error from $15.64\%$ to $3.82\%$, close to the offline-DI Jacobian error $3.64\%$.

\subsubsection{Navier--Stokes.}
\label{sec:navier_stokes_control}
\noindent\textbf{Setup.}
This benchmark is a two-dimensional incompressible flow problem on $\Omega=(0,1)^2$. The distributed forcing $u=(u_1,u_2)$ drives the system
\[
\begin{aligned}
-\mu\Delta y+(y\cdot\nabla)y+\nabla p&=u &&\text{in }\Omega,\\
\nabla\cdot y&=0 &&\text{in }\Omega,\\
y&=0 &&\text{on }\partial\Omega,
\end{aligned}
\qquad \mu=0.1,
\]
with the pressure gauge $\int_\Omega p\,dx=0$. The learned map is $S:u\mapsto(y,p)$, while the function-value and Jacobian metrics in \cref{tab:master_5pde,tab:jac_err} use the velocity component; pressure enters the data loss with weight $0.05$. Forcings are sampled from a solenoidal stream-function prior: for $k,l=1,\ldots,6$, we draw $\beta_{kl}=\xi_{kl}(k^2+l^2)^{-3/2}$ with i.i.d.\ $\xi_{kl}\sim\mathcal N(0,1)$, form $\psi_u=\sum_{k,l}\beta_{kl}\sin^2(k\pi x_1)\sin^2(l\pi x_2)$, and set $u=A_u\nabla^\perp\psi_u/\|\nabla^\perp\psi_u\|_\infty$ with $A_u\sim\mathcal U(16,24)$. The $64\times64$ centered finite-difference forward solver uses a stabilized Stokes initial guess followed by Newton iteration. To remove the collocated pressure checkerboard mode, the discrete continuity row is stabilized as $D y-\alpha h_xh_y Lp=0$ with $\alpha=10^{-3}$; saved pressure fields are mean-subtracted.

\noindent\textbf{Tangent equation.}
For a forcing perturbation $s=\delta u$, the velocity and pressure tangents $(w,\pi)=DS(u)[s]$ satisfy the Oseen saddle system
\[
\begin{aligned}
-\mu\Delta w+(y\cdot\nabla)w+(w\cdot\nabla)y+\nabla\pi&=s,\\
\nabla\cdot w&=0,\\
w|_{\partial\Omega}&=0.
\end{aligned}
\]
Offline DI solves the corresponding stabilized discrete saddle system. The pressure tangent is required by the solve but only $w$ is stored and supervised. Offline DI, sTCL, and evaluation use the same normalized solenoidal direction prior; offline DI stores $200$ training directions, whereas sTCL draws fresh directions on the fly.

\noindent\textbf{Conditioning.}
The conditioning issue has two distinct parts. First, the frozen-state momentum operator is non-self-adjoint and mixes the diffusive scale $\mu|k|^2$ with an advective scale. Second, $(w,\pi)$ form a saddle system: pressure is a Lagrange multiplier, and its gradient occupies the curl-free part of the momentum residual rather than the divergence-free velocity-response subspace. Squaring the unconditioned joint residual therefore combines poorly matched momentum, pressure, continuity, and boundary scales.

The loss handles the momentum block in two steps. It first applies an approximate Leray projector to remove the Fourier gradient component, and then weights the projected residual by an inverse Oseen-symbol magnitude. For a Fourier mode $k$, we use
\[
P(k)=I-\frac{kk^\top}{|k|^2},\qquad
\sigma(k)=\sqrt{(\mu|k|^2)^2+b_0^2|k|^2},\qquad
M_O(k)=\bigl(\sigma(k)+\tau\sigma_{\max}\bigr)^{-1},
\]
with $P(0)=I$ for the mean mode, $\sigma_{\max}=\max_k\sigma(k)$, characteristic speed $b_0=3$, and shift $\tau=0.005$. Because the state-dependent convection field has no single global direction, we use the isotropic magnitude $b_0|k|$. Both $P$ and $M_O$ are applied by a periodic FFT to the interior momentum residual. This is an inexpensive constant-coefficient approximation to the inverse magnitude of the bounded-domain, state-dependent Oseen operator, not a tangent solve.

\noindent\textbf{sTCL form.}
For $w_\theta=Dy_\theta(u)[s]$ and $\pi_\theta=Dp_\theta(u)[s]$, sTCL differentiates the momentum, continuity, and boundary residual blocks:
\[
\begin{aligned}
r_{\mathrm{mom}}&=-\mu\Delta w_\theta+(y_\theta\cdot\nabla)w_\theta
 +(w_\theta\cdot\nabla)y_\theta+\nabla\pi_\theta-s,\\
r_{\mathrm{div}}&=\nabla\cdot w_\theta,
\qquad r_{\mathrm{bc}}=w_\theta|_{\partial\Omega},
\end{aligned}
\]
Writing $(P r_{\mathrm{mom}})_j$ for the $j$th Cartesian component of the projected momentum residual, with $j\in\{1,2\}$, the momentum energy is
\[
\mathcal L_{\mathrm{mom}}
=\sum_{j=1}^2\left\langle (P r_{\mathrm{mom}})_j,
M_O(P r_{\mathrm{mom}})_j\right\rangle,
\]
and the loss combines the three blocks as
\[
\mathcal L_{\mathrm{tan}}
=\mathcal L_{\mathrm{mom}}
+0.05\,\operatorname{mean}(r_{\mathrm{div}}^2)
+\operatorname{mean}(r_{\mathrm{bc}}^2),
\]
before the common derivative/data EMA rescaling. The separate continuity and boundary terms are essential: projecting the momentum equation does not enforce either $\nabla\cdot w_\theta=0$ or the no-slip tangent condition, and the periodic projector is only approximate on a bounded Dirichlet domain. The on-the-fly continuity block uses the raw discrete divergence rather than the pressure-stabilized row used by the collocated reference solver. At $N{=}512$, this loss reduces velocity error from $0.619\%$ to $0.400\%$ and velocity-JVP error from $1.084\%$ to $0.570\%$ without tangent labels; at $N{=}2048$, the corresponding errors are $0.198\%$ and $0.282\%$.

\subsection{Per-cell hyperparameter settings}
\label{app:per_cell_configs}

\Cref{tab:per_cell_configs} lists the hyperparameter setting backing every cell of \cref{tab:master_5pde} and \cref{tab:jac_err}. All derivative-informed cells use the EMA-normalized weighting of \cref{sec:method_stcl} (decay $0.99$ for both running averages), so $\lambda$ is the target derivative-to-data loss ratio for either method. Burgers and Allen--Cahn use PDE-level validation grids, $\lambda\in\{0.1,0.25,0.5,1.0,2.0\}$ and $\lambda\in\{0.1,0.3,0.5,1.0,2.0\}$, which select the common values $\lambda=1$ and $\lambda=2$, respectively. Each value is then shared across training sizes and both derivative-informed methods. For nonlinear diffusion, each $(N,\mathrm{method})$ is selected by validation from $\lambda\in\{0.1,0.3,0.5,0.7,0.9,1.0\}$; all reported cells select $\lambda=1$. Helmholtz uses per-cell selection from $\lambda\in\{0.1,0.3,0.5,0.7,0.9\}$. Navier--Stokes uses per-cell selection from $\lambda\in\{0.1,0.25,0.5,1.0\}$. Its values are selected by mean validation velocity error across seeds, and the same selected checkpoints supply both tables. The learning rate is fixed at $\eta=10^{-3}$, and we do not perform a broader joint search over $(\eta,\lambda)$. The sTCL column uses the operator-specific form from \cref{tab:benchmark_conditioning}: a diffusivity-scaled DST inverse-Laplacian residual for nonlinear diffusion, MINRES-$5$ target matching for Helmholtz, an RHS-normalized space--time residual for Burgers, a raw space--time $L^2$ residual for Allen--Cahn, and a Leray--Oseen-preconditioned saddle residual for Navier--Stokes.

\begin{table}[ht]
\apptablesetup
\caption{Per-cell hyperparameter settings for \cref{tab:master_5pde} and \cref{tab:jac_err}. All cells use Adam at $\eta=10^{-3}$, batch size $32$, and three seeds $\{0,1,2\}$. All derivative-informed cells use the common EMA-normalized weighting. Both derivative-informed methods use $q=4$ directions per training example at each update, drawn from matched PDE-specific distributions; offline DI draws direction--label pairs from a stored bank of PDE-dependent size $r$, whereas sTCL uses unlabeled directions.}
\label{tab:per_cell_configs}
\resizebox{\linewidth}{!}{%
\begin{tabular}{l c c c l}
\toprule
PDE & $\lambda$ & Offline-DI bank $r$ & sTCL form & Notes \\
\midrule
Helmholtz                & per-$N$ (see below) & $289$ & MINRES-$5$ target-matching & validation-selected 5-pt $\lambda$ grid \\
\multicolumn{5}{l}{\quad $N{=}128$: $\lambda{=}0.1$ for both methods; $N{=}512$: $\lambda{=}0.5$ for both; $N{=}1024$: $\lambda{=}0.9$ (offline DI), $\lambda{=}0.7$ (sTCL).} \\
Nonlin.\ diffusion       & $1$   & $289$ & DST inverse-Laplacian residual & per-cell selected from 6-pt grid \\
Burgers             & $1$   & $200$ & RHS-normalized space--time residual & PDE-level selected from 5-pt grid \\
Allen--Cahn         & $2$   & $200$ & raw space--time $L^2$ residual & PDE-level selected from 5-pt grid \\
Navier--Stokes   & per-$N$ (see below) & $200$ & Leray--Oseen-preconditioned residual & validation-selected 4-pt $\lambda$ grid \\
\multicolumn{5}{l}{\quad $N{=}512$: $\lambda{=}0.5$ (offline DI), $0.25$ (sTCL); $N{=}1024$: $1$ (offline DI), $0.25$ (sTCL); $N{=}2048$: $0.5$ for both.} \\
\bottomrule
\end{tabular}
}
\end{table}

\subsection{Training time and offline tangent-data generation cost}
\label{app:offline_cost}

\Cref{tab:wallclock_per_pde} reports the per-job wall-clock time at $N{=}512$ under each benchmark's full training budget: $2000$ epochs for Helmholtz, nonlinear diffusion, and Burgers, $1500$ epochs for Allen--Cahn, and $5000$ epochs for Navier--Stokes. sTCL and offline DI have training costs of the same order.

The offline-DI training times in \cref{tab:wallclock_per_pde} exclude the generation of the tangent labels, which is the stage that sTCL removes; \cref{tab:offline_cost} reports its cost. For each benchmark, we time the tangent solver that produces the offline-DI labels with a single thread on one core of an Intel Xeon Platinum~8559C. Helmholtz and nonlinear diffusion--reaction use sparse direct solves of the linearized finite-difference system; Burgers and Allen--Cahn use forward-mode differentiation through the implicit-Picard and Fourier-spectral IMEX solvers, respectively; and Navier--Stokes uses a sparse direct solve of the stabilized Oseen saddle system.

\begin{table}[t]
\centering
\small
\caption{Single-core CPU cost of offline tangent-label generation. The unit cost is the time per sample and direction, and the total is the unit cost multiplied by the stored direction count $r$ and the largest training size $N$. sTCL does not require this stage.}
\label{tab:offline_cost}
\begin{tabular}{l r r r r}
\toprule
PDE & Unit cost (ms) & $r$ & Max $N$ & Total (CPU-hours) \\
\midrule
Helmholtz          & $2.0$  & $289$ & $1024$ & $0.16$ \\
Nonlin.\ diffusion & $8.8$  & $289$ & $1024$ & $0.72$ \\
Burgers            & $460$  & $200$ & $512$  & $13.1$ \\
Allen--Cahn        & $115$  & $200$ & $2048$ & $13.1$ \\
Navier--Stokes          & $90.1$ & $200$ & $2048$ & $10.3$ \\
\bottomrule
\end{tabular}
\end{table}

The offline cost is small for the two elliptic benchmarks, about $10$ and $43$ CPU-minutes, but reaches about $13$ CPU-hours for Burgers and Allen--Cahn, where each label requires differentiating through a full time integration, and $10.3$ CPU-hours for Navier--Stokes. Tangent solves for different samples are independent, so parallelizing over CPU cores reduces the wall-clock time but not the total CPU time. This cost scales with $N r$, and the labels must be regenerated whenever the input distribution, resolution, or direction bank changes; the per-step cost of sTCL, by contrast, depends on $q$ but not on $N r$. Storage also scales with $N r$: at its largest training size, $N{=}1024$, the nonlinear-diffusion label bank occupies about $5$~GB in single precision.

\subsection{MINRES inner-iteration count}
\label{app:minres_iters}

\Cref{tab:minres_iters} examines how accurately the on-the-fly MINRES target must be constructed for Helmholtz. At $N{=}512$, we vary only the inner-iteration count $m$ while holding the training protocol fixed at $2000$ epochs, lr$=10^{-3}$, and $\lambda=0.1$ (the main $N{=}512$ comparison uses the validation-selected $\lambda=0.5$). Keeping $\lambda$ fixed ensures that the differences in \cref{tab:minres_iters} reflect the MINRES iteration count alone. Most of the improvement occurs within the first few iterations: increasing $m$ from $1$ to $3$ reduces the function-value error from $3.50\%$ to $2.52\%$ and the Jacobian error from $23.03\%$ to $15.52\%$. Beyond $m=5$, both errors change only slightly relative to seed variability, while training time continues to increase. Thus $m=5$ provides a practical cost--accuracy balance and confirms that the on-the-fly target need only be accurate enough to condition the loss, rather than reproduce a high-accuracy offline tangent solve.

\begin{table}[t]
\centering
\small
\caption{Effect of the MINRES inner-iteration count $m$ on Helmholtz at $N{=}512$ with a $2000$-epoch training budget and fixed $\lambda=0.1$; the main $N{=}512$ comparison in \cref{tab:master_5pde,tab:jac_err} uses the validation-selected $\lambda=0.5$. Errors are reported as mean $\pm$ standard deviation over three seeds.}
\label{tab:minres_iters}
\setlength{\tabcolsep}{5pt}
\begin{tabular}{r r r r}
\toprule
$m$ & Function rel.\ err. & Jacobian rel.\ err. & Training time (s) \\
\midrule
$\phantom{0}1$ & $\phantom{0}3.50 \pm 0.45\%$ & $23.03 \pm 2.52\%$ & $2056$ \\
$\phantom{0}2$ & $\phantom{0}2.80 \pm 0.20\%$ & $18.32 \pm 1.48\%$ & $2132$ \\
$\phantom{0}3$ & $\phantom{0}2.52 \pm 0.22\%$ & $15.52 \pm 1.56\%$ & $2228$ \\
$\phantom{0}5$ & $\phantom{0}2.47 \pm 0.28\%$ & $14.71 \pm 2.09\%$ & $2412$ \\
$10$           & $\phantom{0}2.46 \pm 0.21\%$ & $14.61 \pm 1.65\%$ & $2884$ \\
$25$           & $\phantom{0}2.41 \pm 0.15\%$ & $14.51 \pm 1.22\%$ & $4216$ \\
\bottomrule
\end{tabular}
\end{table}

\subsection{Sketch count, resolution, and gradient variance}
\label{app:q_scaling}

The sketch count $q$ controls a cost--variance tradeoff. Each direction costs one forward-mode JVP regardless of the input dimension, but the variance of the sketch estimator can depend on the effective rank and sensitivity spectrum of the residual operator, so we test whether $q$ must grow with resolution to keep sTCL useful. With the normalization in \eqref{eq:empirical_stcl}, every $q\geq1$ estimates the same expected sTCL objective and gradient; increasing $q$ averages more independent directions at an optimizer step. Because the directions change from step to step, $q$ is not the total number of directions observed over training. We examine $q\in\{0,1,2,4,8,16,32\}$, with $q=0$ denoting solution-only training, using $N=128$ training samples, $2000$ epochs, and $\lambda=1$. The nonlinear-diffusion sweep reports mean and sample standard deviation over training seeds $\{0,1,2\}$.

\noindent\textbf{Accuracy versus resolution.}
\Cref{tab:q_resolution_accuracy} reports nonlinear-diffusion results on $32^2$, $64^2$, and $128^2$ grids, corresponding to input dimensions $1024$, $4096$, and $16{,}384$. A single sketch direction already reduces both errors substantially at every resolution. Relative to the three-seed $q=0$ means, $q=1$ reduces field error by $50.7\%$, $49.3\%$, and $36.9\%$, and JVP error by $44.8\%$, $43.5\%$, and $32.1\%$, respectively. Among $q\geq1$, neither metric improves monotonically with $q$: the lowest mean field and JVP errors occur at $q=4$ on $32^2$ and at $q=16$ on both larger grids. Both errors grow moderately with resolution for every $q\geq1$ (field errors of $4.18$--$4.41\%$ on $32^2$ versus $5.53$--$5.88\%$ on $128^2$), and increasing $q$ from $1$ to $32$ does not remove this growth. The resolution dependence is therefore not caused by insufficient sketch averaging; resolution is entangled with other factors, such as a harder discretized learning problem at fixed model capacity, training-set size, and training budget.

\begin{table}[ht]
\centering\footnotesize
\setlength{\tabcolsep}{2.5pt}
\renewcommand{\arraystretch}{1.04}
\caption{Nonlinear-diffusion accuracy versus sketch count and spatial resolution. Each entry is the mean $\pm$ sample standard deviation of the relative error (\%) over three training seeds $\{0,1,2\}$. For each grid, the paired columns report field and JVP errors.}
\label{tab:q_resolution_accuracy}
\begin{tabular}{c r r r r r r}
\toprule
$q$ & \multicolumn{2}{c}{$32^2$} & \multicolumn{2}{c}{$64^2$} & \multicolumn{2}{c}{$128^2$} \\
\cmidrule(lr){2-3}\cmidrule(lr){4-5}\cmidrule(lr){6-7}
& Field & JVP & Field & JVP & Field & JVP \\
\midrule
$0$  & $8.94\pm0.26$ & $28.53\pm0.53$ & $9.19\pm0.32$ & $29.33\pm0.68$ & $9.33\pm0.28$ & $29.84\pm0.62$ \\
$1$  & $4.41\pm0.23$ & $15.74\pm0.82$ & $4.66\pm0.05$ & $16.56\pm0.45$ & $5.88\pm0.44$ & $20.25\pm1.06$ \\
$2$  & $4.33\pm0.21$ & $15.46\pm0.86$ & $4.65\pm0.29$ & $16.70\pm0.75$ & $5.66\pm0.13$ & $19.70\pm0.26$ \\
$4$  & $4.18\pm0.34$ & $15.02\pm1.05$ & $4.65\pm0.11$ & $16.51\pm0.39$ & $5.88\pm0.09$ & $19.91\pm0.33$ \\
$8$  & $4.27\pm0.54$ & $15.56\pm1.74$ & $4.57\pm0.29$ & $16.26\pm0.85$ & $5.72\pm0.15$ & $19.65\pm0.51$ \\
$16$ & $4.37\pm0.35$ & $15.74\pm1.22$ & $4.50\pm0.15$ & $16.24\pm0.50$ & $5.53\pm0.13$ & $19.35\pm0.20$ \\
$32$ & $4.31\pm0.26$ & $15.44\pm0.94$ & $4.64\pm0.29$ & $16.61\pm0.59$ & $5.66\pm0.05$ & $19.69\pm0.03$ \\
\bottomrule
\end{tabular}
\end{table}

In a separate $N{=}128$ runtime sweep on the $64^2$ grid, the wall time grows approximately linearly with $q$ (\cref{tab:q_runtime}).

\begin{table}[ht]
\apptablesetup
\caption{Single-run nonlinear-diffusion training time (minutes) in the controlled $N{=}128$ sketch-count sweep on the $64^2$ grid. These profiling values are not across-seed averages.}
\label{tab:q_runtime}
\begin{tabular}{l r r r r r r r}
\toprule
Benchmark & $q=0$ & $q=1$ & $q=2$ & $q=4$ & $q=8$ & $q=16$ & $q=32$ \\
\midrule
Nonlin. diffusion & $1.95$ & $3.69$ & $5.53$ & $8.46$ & $15.05$ & $29.23$ & $54.61$ \\
\bottomrule
\end{tabular}
\end{table}

\noindent\textbf{Gradient-variance measurement.}
For optimizer step $i=1,\ldots,n_{\mathrm b}$ in epoch $e$, where $n_{\mathrm b}$ is the number of optimizer steps per epoch, let $g_{e,i}^{\mathrm{data}}$, $g_{e,i}^{\mathrm{raw}}$, and
$g_{e,i}^{\mathrm{weighted}}=\operatorname{sg}(\gamma_{e,i})g_{e,i}^{\mathrm{raw}}$ denote the data, raw-sTCL, and EMA-weighted-sTCL gradients, where $\gamma_{e,i}$ is the EMA coefficient $\gamma_t$ of \cref{sec:method_stcl} at that step and $\operatorname{sg}(\cdot)$ denotes stop-gradient. Before clipping,
$g_{e,i}^{\mathrm{total}}=g_{e,i}^{\mathrm{data}}+g_{e,i}^{\mathrm{weighted}}$. For each component $a\in\{\mathrm{raw},\mathrm{weighted},\mathrm{total}\}$, we measure the within-epoch trace covariance
\[
V_a^{(e)}=\frac{1}{n_{\mathrm b}-1}\sum_{i=1}^{n_{\mathrm b}}
\left\|g_{e,i}^{a}-\bar g_e^a\right\|_2^2,
\qquad
\bar g_e^a=\frac{1}{n_{\mathrm b}}\sum_{i=1}^{n_{\mathrm b}}g_{e,i}^{a}.
\]
For each training seed, we average $V_a^{(e)}$ over epochs $26$--$2000$, $\overline V_a=\tfrac{1}{1975}\sum_{e=26}^{2000}V_a^{(e)}$, excluding the initial transient. We then report the mean and sample standard deviation of $\overline V_a$ across seeds. Thus the within-epoch trace variance and its across-seed variability are distinct quantities; the former is measured over the $n_{\mathrm b}=N/B=4$ changing optimizer steps per epoch and is not a frozen-model estimate of sketch-only randomness.

\begin{table}[H]
\centering\footnotesize
\setlength{\tabcolsep}{2.5pt}
\renewcommand{\arraystretch}{1.04}
\caption{Gradient trace variance $\overline V_a$ for nonlinear diffusion, averaged over epochs $26$--$2000$ and reported as mean $\pm$ sample standard deviation over three training seeds. The Raw sTCL, Weighted sTCL, and Total columns report $\overline V_{\mathrm{raw}}$, $\overline V_{\mathrm{weighted}}$, and $\overline V_{\mathrm{total}}$, respectively.}
\label{tab:q_gradient_variance}
\begin{tabular}{c r r r r r}
\toprule
Grid & Dim. & $q$ & Raw sTCL & Weighted sTCL & Total \\
\midrule
$32^2$ & $1024$ & $0$  & $0$ & $0$ & $\msdexp{8.710}{1.520}{-7}$ \\
$32^2$ & $1024$ & $1$  & $\msdexp{1.043}{0.297}{-2}$ & $\msdexp{9.832}{5.880}{-7}$ & $\msdexp{6.765}{3.170}{-6}$ \\
$32^2$ & $1024$ & $2$  & $\msdexp{4.509}{0.311}{-3}$ & $\msdexp{4.327}{0.970}{-7}$ & $\msdexp{5.936}{1.560}{-6}$ \\
$32^2$ & $1024$ & $4$  & $\msdexp{2.578}{0.429}{-3}$ & $\msdexp{2.627}{0.830}{-7}$ & $\msdexp{5.712}{2.700}{-6}$ \\
$32^2$ & $1024$ & $8$  & $\msdexp{1.272}{0.125}{-3}$ & $\msdexp{1.202}{0.279}{-7}$ & $\msdexp{4.726}{2.400}{-6}$ \\
$32^2$ & $1024$ & $16$ & $\msdexp{7.590}{1.300}{-4}$ & $\msdexp{7.163}{2.410}{-8}$ & $\msdexp{4.294}{1.960}{-6}$ \\
$32^2$ & $1024$ & $32$ & $\msdexp{5.404}{1.300}{-4}$ & $\msdexp{5.651}{2.760}{-8}$ & $\msdexp{5.120}{3.180}{-6}$ \\
\midrule
$64^2$ & $4096$ & $0$  & $0$ & $0$ & $\msdexp{7.000}{0.686}{-7}$ \\
$64^2$ & $4096$ & $1$  & $\msdexp{2.359}{0.843}{-2}$ & $\msdexp{4.628}{3.010}{-7}$ & $\msdexp{7.051}{2.080}{-6}$ \\
$64^2$ & $4096$ & $2$  & $\msdexp{1.126}{0.224}{-2}$ & $\msdexp{2.294}{0.894}{-7}$ & $\msdexp{5.512}{1.890}{-6}$ \\
$64^2$ & $4096$ & $4$  & $\msdexp{7.781}{2.330}{-3}$ & $\msdexp{1.756}{1.070}{-7}$ & $\msdexp{4.712}{0.737}{-6}$ \\
$64^2$ & $4096$ & $8$  & $\msdexp{3.660}{0.854}{-3}$ & $\msdexp{7.022}{2.020}{-8}$ & $\msdexp{4.758}{1.810}{-6}$ \\
$64^2$ & $4096$ & $16$ & $\msdexp{2.259}{0.786}{-3}$ & $\msdexp{4.432}{2.430}{-8}$ & $\msdexp{5.343}{1.270}{-6}$ \\
$64^2$ & $4096$ & $32$ & $\msdexp{1.803}{0.951}{-3}$ & $\msdexp{3.598}{2.280}{-8}$ & $\msdexp{4.436}{2.480}{-6}$ \\
\midrule
$128^2$ & $16{,}384$ & $0$  & $0$ & $0$ & $\msdexp{7.600}{2.540}{-7}$ \\
$128^2$ & $16{,}384$ & $1$  & $\msdexp{1.010}{0.356}{-1}$ & $\msdexp{3.567}{2.870}{-7}$ & $\msdexp{5.399}{0.668}{-6}$ \\
$128^2$ & $16{,}384$ & $2$  & $\msdexp{4.398}{1.120}{-2}$ & $\msdexp{1.640}{0.904}{-7}$ & $\msdexp{3.956}{1.010}{-6}$ \\
$128^2$ & $16{,}384$ & $4$  & $\msdexp{3.651}{0.580}{-2}$ & $\msdexp{1.417}{0.528}{-7}$ & $\msdexp{4.343}{1.040}{-6}$ \\
$128^2$ & $16{,}384$ & $8$  & $\msdexp{1.475}{0.232}{-2}$ & $\msdexp{5.379}{1.730}{-8}$ & $\msdexp{3.484}{0.696}{-6}$ \\
$128^2$ & $16{,}384$ & $16$ & $\msdexp{8.628}{4.470}{-3}$ & $\msdexp{3.321}{2.390}{-8}$ & $\msdexp{3.639}{1.510}{-6}$ \\
$128^2$ & $16{,}384$ & $32$ & $\msdexp{6.479}{3.610}{-3}$ & $\msdexp{2.459}{2.110}{-8}$ & $\msdexp{3.628}{1.810}{-6}$ \\
\bottomrule
\end{tabular}
\end{table}

At fixed resolution, $\overline V_{\mathrm{raw}}$ decreases as $q$ grows: from $q=1$ to $q=32$, its mean falls by $94.8\%$, $92.4\%$, and $93.6\%$ on the $32^2$, $64^2$, and $128^2$ grids, and the mean $\overline V_{\mathrm{weighted}}$ falls by $94.3\%$, $92.2\%$, and $93.1\%$. The optimizer, however, sees the weighted term, and $\overline V_{\mathrm{weighted}}$ is only a small fraction of $\overline V_{\mathrm{total}}$: $0.7$--$14.5\%$ across all grids and $q\geq1$, and $3.3$--$4.6\%$ at $q=4$. Increasing $q$ therefore mainly reduces an already small part of the gradient noise and, since the expected objective does not depend on $q$, need not improve the final accuracy (\cref{tab:q_resolution_accuracy}). The mean $\overline V_{\mathrm{total}}$ does fall from $q=1$ to $q=32$, by $24.3\%$, $37.1\%$, and $32.8\%$, but not monotonically in $q$ and by more than the weighted share of $\overline V_{\mathrm{total}}$, so this change cannot come from the reduced sTCL variance alone. The picture is the same across resolutions: at fixed $q\geq1$, $\overline V_{\mathrm{raw}}$ increases with resolution, but $\overline V_{\mathrm{weighted}}$ decreases, and $\overline V_{\mathrm{total}}$ on $128^2$ is below its $32^2$ value for every $q\geq1$. Thus $q$ did not need to grow with resolution: a fixed small $q$ remained effective over the tested $16$-fold increase in grid dimension. This is consistent with the structure of the inputs: at every grid, the inputs and the probes are drawn from the same $800$-mode cosine KLE, so they lie on the same input manifold of dimension at most $800$ while the grid dimension grows from $1024$ to $16{,}384$, which suggests that the required $q$ is set by the effective dimension and sensitivity spectrum of the input distribution rather than by the number of grid points. We therefore treat $q=4$ as a conservative empirical operating point rather than a dimension-independent guarantee: start with a small $q$ and increase it only when $\overline V_{\mathrm{weighted}}$ becomes a substantial fraction of $\overline V_{\mathrm{total}}$ or validation accuracy is sensitive to $q$. These sweeps do not isolate resolution from the other factors above and do not establish a general scaling law.

\subsection{Long-run convergence check}
\label{app:long_run_convergence}

We examine representative long-run settings for all five main benchmarks while keeping the architectures, loss weights, optimizers, and tangent-direction sampling protocols unchanged. The selected settings are Burgers with $N{=}512$, Helmholtz with $N{=}1024$, nonlinear diffusion with $N{=}1024$, Allen--Cahn with $N{=}1024$, and Navier--Stokes with $N{=}2048$. The first four are run for $4000$ epochs, and Navier--Stokes is run for $5000$ epochs; its validation-selected offline-DI and sTCL configurations both use $\lambda=0.5$.

\begin{figure}[!t]
\centering
\begin{subfigure}{0.48\linewidth}
\centering
\includegraphics[width=\linewidth]{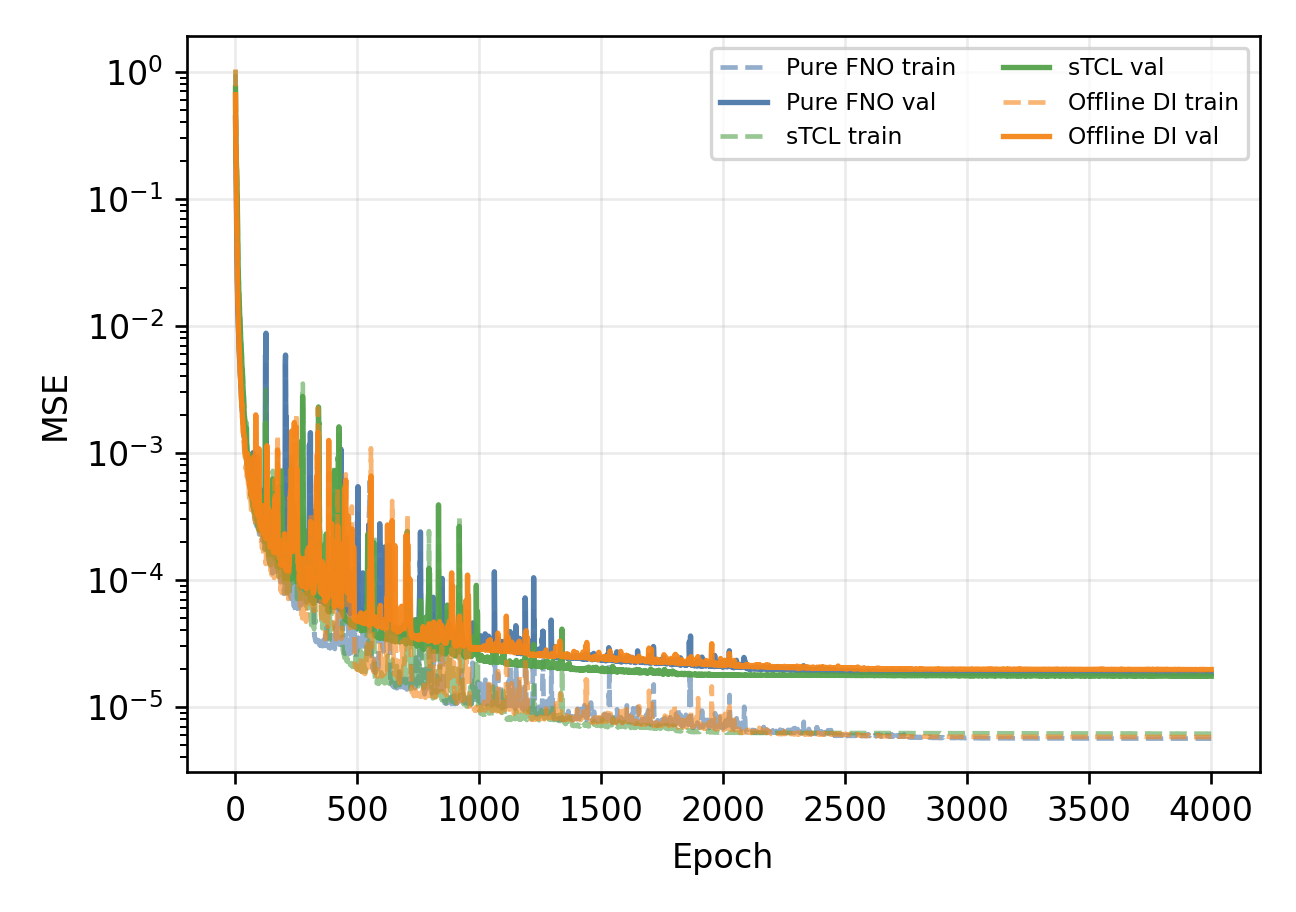}
\caption{Burgers, $N{=}512$.}
\end{subfigure}
\hfill
\begin{subfigure}{0.48\linewidth}
\centering
\includegraphics[width=\linewidth]{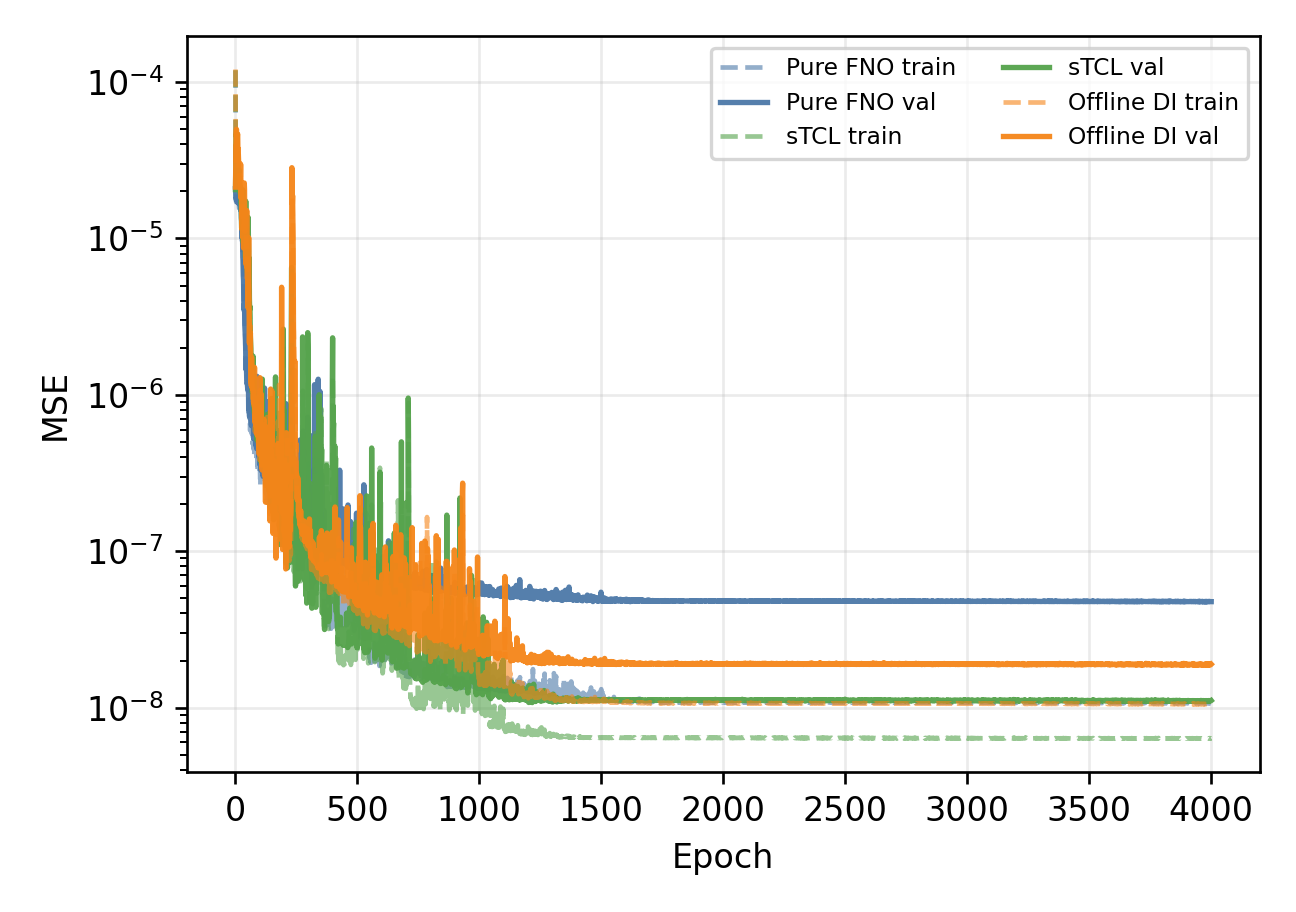}
\caption{Helmholtz, $N{=}1024$.}
\end{subfigure}

\vspace{0.4em}
\begin{subfigure}{0.48\linewidth}
\centering
\includegraphics[width=\linewidth]{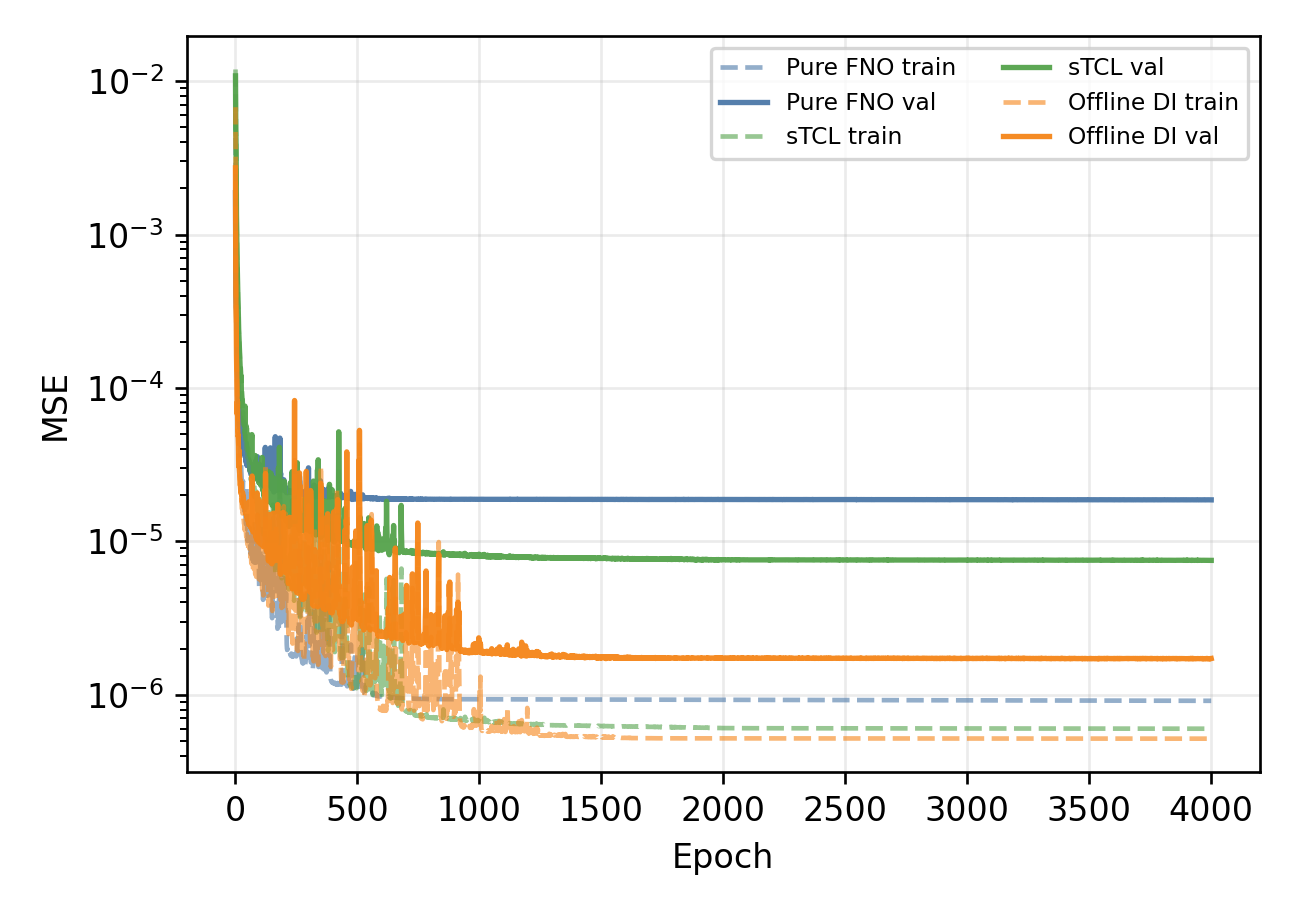}
\caption{Nonlinear diffusion, $N{=}1024$.}
\end{subfigure}
\hfill
\begin{subfigure}{0.48\linewidth}
\centering
\includegraphics[width=\linewidth]{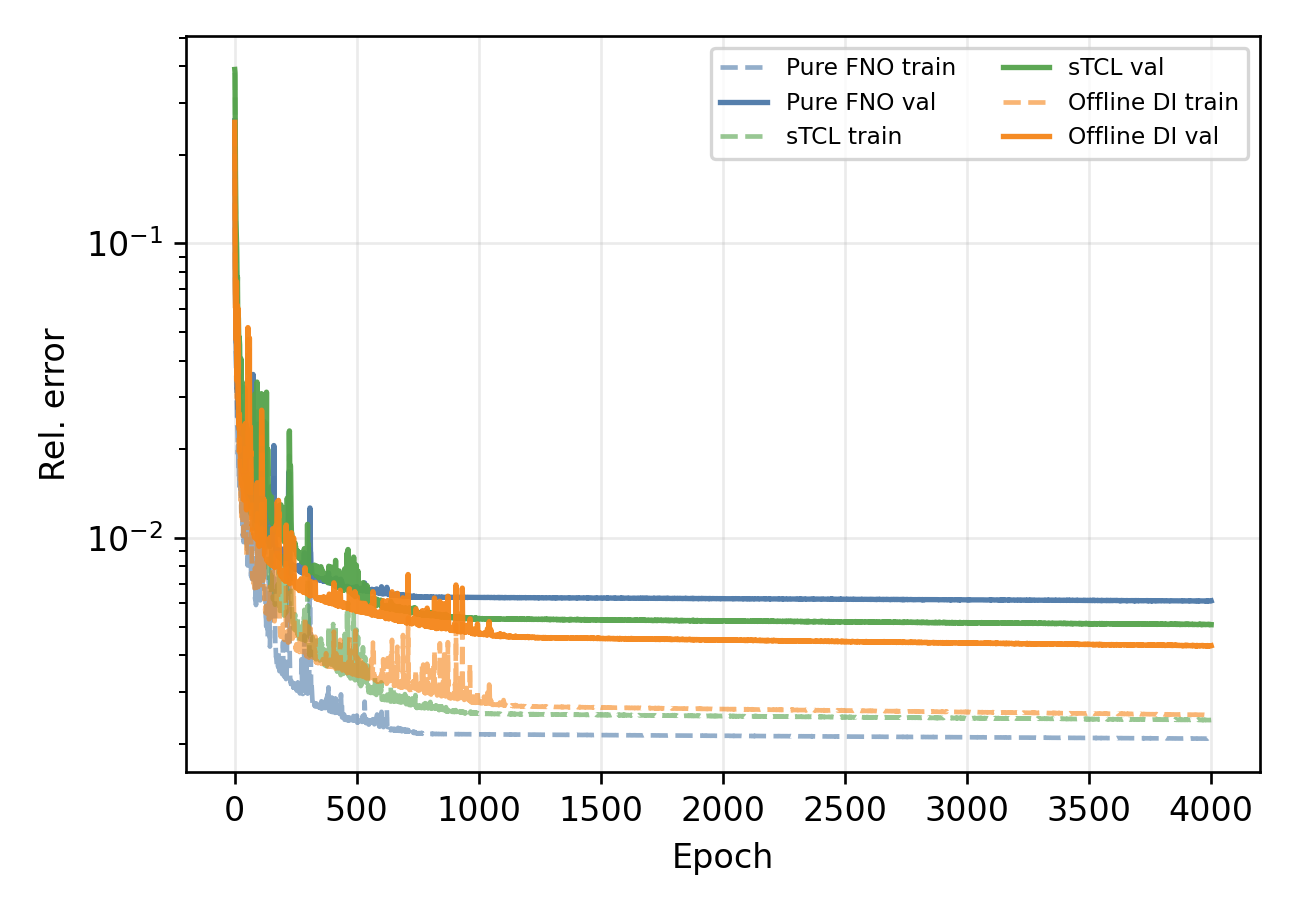}
\caption{Allen--Cahn, $N{=}1024$.}
\end{subfigure}

\vspace{0.4em}
\begin{subfigure}{0.48\linewidth}
\centering
\includegraphics[width=\linewidth]{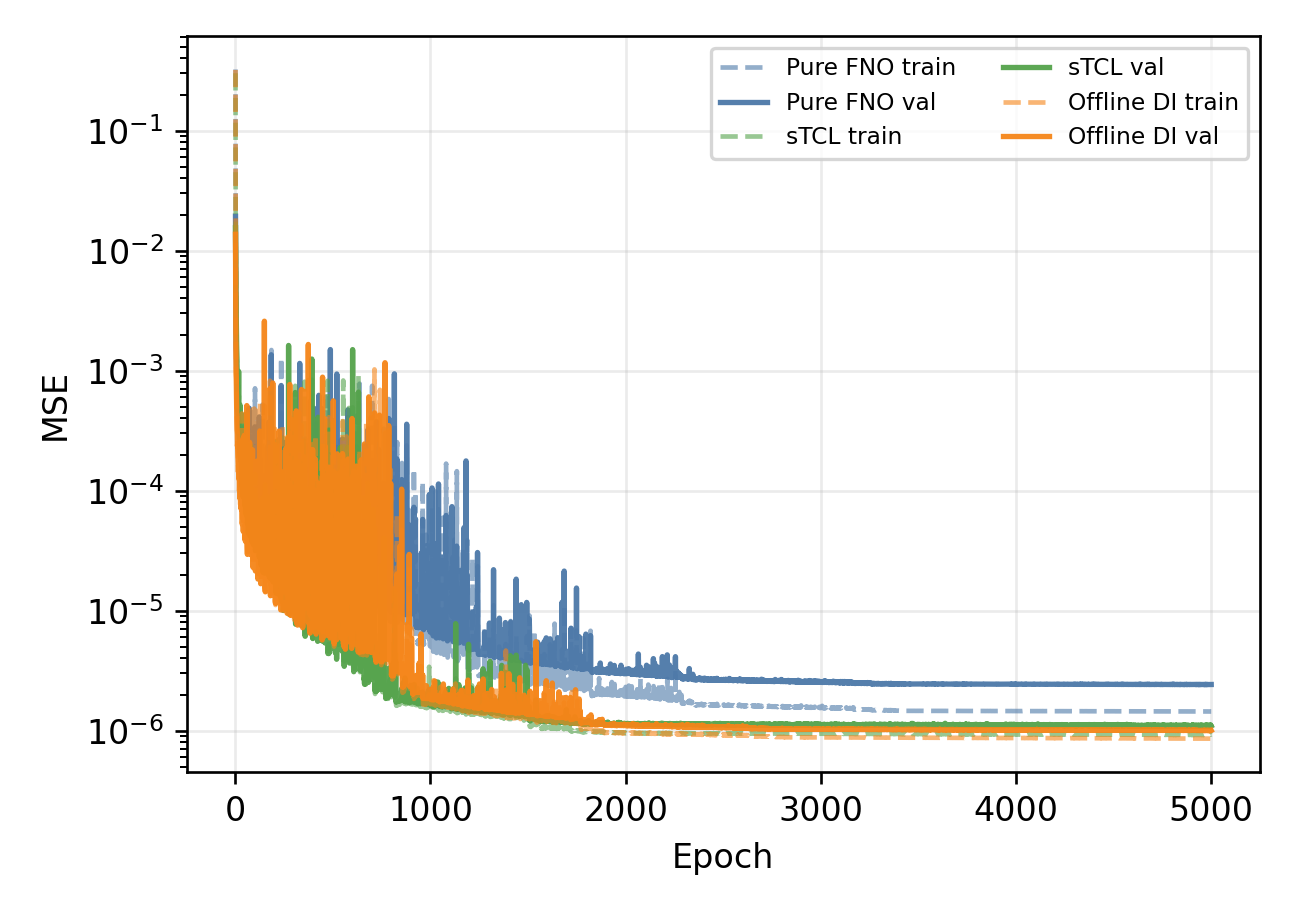}
\caption{Navier--Stokes, $N{=}2048$.}
\end{subfigure}
\caption{Single-seed training and validation trajectories for representative benchmark settings: $4000$ epochs for Burgers, Helmholtz, nonlinear diffusion, and Allen--Cahn, and $5000$ epochs for Navier--Stokes. Panels (a)--(c) and (e) show the MSE, and panel (d) shows the relative error. Dashed curves denote training and solid curves denote validation for pure FNO, sTCL, and offline DI; these curves are convergence diagnostics rather than across-seed uncertainty summaries.}
\label{fig:long_run_convergence}
\end{figure}

\Cref{fig:long_run_convergence} shows that the validation curves for pure FNO, offline DI, and sTCL stabilize within these long-run budgets, with only small late-stage fluctuations and no change in the qualitative ordering of the methods. These trajectories indicate that the reported results reflect the converged behavior of the training objectives rather than transient optimization effects.

\section{Discussion and practical recommendations}
\label{sec:discussion}

\paragraph{Practical decision rule.}
The five-PDE study suggests the following rule:
\begin{enumerate}[nosep,leftmargin=*]
    \item \emph{What type of tangent operator does the PDE have?} Use the operator structure to select a practical loss form from \cref{tab:benchmark_conditioning}: unpreconditioned space--time residuals for the parabolic problems considered here (Burgers and Allen--Cahn), a diffusivity-scaled DST inverse-Laplacian residual for SPD elliptic problems (nonlinear diffusion), MINRES-$m$ target matching for symmetric indefinite problems (Helmholtz), and Leray--Oseen-preconditioned residuals for incompressible non-self-adjoint saddle systems, as in Navier--Stokes.
    \item \emph{Offline wall-clock budget?} For small problems where offline tangent generation is cheap, offline DI and sTCL are competitive. For larger problems where offline cost \emph{dominates}, sTCL becomes attractive because it eliminates that stage.
\end{enumerate}

\paragraph{When sTCL's derivative-label-free advantage matters most.}
The qualitative advantages of sTCL over offline-DI-style methods are greatest when: (i) adding training data after the fact requires rerunning the offline tangent pipeline, (ii) disk storage is constrained, (iii) the number or distribution of sketch directions may change, or (iv) the forward solver is expensive enough that a large offline sensitivity pass is impractical.

\paragraph{Are we just moving offline cost into on-the-fly training?}
A natural objection to on-the-fly sensitivity training is that the offline tangent-solve cost has merely been moved into the inner training loop: if the method applies $A^{-1}$ through short iterative solves at every step, is the per-step preconditioning work just the offline cost amortized over training? In sTCL, it is not. The reason links the conditioning analysis (\cref{sec:method_conditioning}) to the surrogate's training dynamics: \emph{we do not need an optimal tangent solve, only a solve accurate enough for the surrogate's current accuracy stage}. Offline DI precomputes $J\cdot v$ to solver accuracy because the labels must drive training at every later stage. By contrast, in sTCL the residual or MINRES target is constructed \emph{on the fly}, so its accuracy requirement scales with the loss itself: at the start of training, when $\|S_\theta(f) - S(f)\|$ is tens of percent, even a one-iteration tangent solve produces a useful descent direction; later in training, as the surrogate improves, the same five MINRES iterations remain adequate because the conditioned derivative signal only needs to correlate with the true Sobolev gradient at the \emph{current} operating point. Empirically (\cref{tab:minres_iters}), increasing the inner count from $m{=}1$ to $m{=}3$ gives the dominant improvement, while $m{=}5$ remains within seed variance of $m{=}25$ on Helmholtz at substantially lower per-step cost. The same ``good-enough'' principle explains why the DST inverse-Laplacian metric, a single fast sine transform and diagonal scaling with no iteration, is sufficient for the SPD elliptic case (\cref{sec:nonlinear_diffusion_difno}): it approximates the principal inverse closely enough on the spectrum populated by the residual. The total preconditioning work is therefore not the offline cost re-amortized; it is the smaller cost of producing descent directions at the accuracy level training needs.

\end{document}